\documentclass{article}

\usepackage[final]{corl_2021} %

\usepackage{graphicx}
\usepackage{caption}
\usepackage{amsmath,amssymb,enumerate}

\usepackage{amsthm}
\usepackage{mathtools}
\usepackage{breqn}
\usepackage{algorithm, algorithmicx, algpseudocode}

\usepackage{blindtext}
\usepackage{gensymb}
\usepackage{xparse}
\usepackage{lipsum}
\usepackage{mathrsfs}
\usepackage[mathscr]{euscript}
\usepackage{times}
\usepackage[noadjust]{cite} %
\usepackage{multicol}
\usepackage[caption=false,font=footnotesize]{subfig}
\usepackage{amsfonts}
\usepackage[T1]{fontenc}
\usepackage{textcomp}
\usepackage{amsfonts}
\usepackage{soul}
\usepackage{multirow}
\usepackage{booktabs}

\newcommand{\squeezeup}{\vspace{-3mm}}

\newcommand{\m}{\mathop{\mathrm{m}}}
\newcommand{\rad}{\mathop{\mathrm{rad}}}

\newcommand{\Hz}{\mathop{\mathrm{Hz}}}

\newcommand{\transpose}{\mathsf{T}}

\title{Legged Robot State Estimation using Invariant Kalman Filtering and Learned Contact Events}

\author{
  Tzu-Yuan Lin, Ray Zhang, Justin Yu, and Maani Ghaffari\\
  University of Michigan, Ann Arbor, MI, USA \\
  \texttt{\{tzuyuan,rzh,yujustin,maanigj\}@umich.edu} 
}

\begin{document}

\maketitle
\begin{abstract}
This work develops a learning-based contact estimator for legged robots that bypasses the need for physical sensors and takes multi-modal proprioceptive sensory data as input. Unlike vision-based state estimators, proprioceptive state estimators are agnostic to perceptually degraded situations such as dark or foggy scenes. While some robots are equipped with dedicated physical sensors to detect necessary contact data for state estimation, some robots do not have dedicated contact sensors, and the addition of such sensors is non-trivial without redesigning the hardware. The trained network can estimate contact events on different terrains. The experiments show that a contact-aided invariant extended Kalman filter can generate accurate odometry trajectories compared to a state-of-the-art visual SLAM system, enabling robust proprioceptive odometry.
\end{abstract}

\section{Introduction}
\label{sec:introduction}
Legged robots can traverse uneven terrains. This unique capability gives them the potential to conduct scientific exploration in extreme environments, execute rescue missions in hazardous scenes, and can help humans in everyday tasks~\citep{raibert1989dynamically}. To accomplish these goals, knowledge of the robot's current state is necessary. While joint angles can be directly measured using encoders, the robot's pose and velocity require additional measurements and a mathematically sound data fusion method.

Vision-based state estimators might suffer from failure due to illumination changes or extreme environments such as snowstorms or foggy scenes~\citep{bijelic2018benchmark}. In contrast, proprioceptive state estimators are agnostic to perceptually degraded situations and can operate at a high update rate (e.g., $100-2000 \Hz$). This high-frequency information provides odometry estimates for localization and mapping tasks~\citep{belter2019optimization,gan2019bayesian,wisth2020preintegrated} and controllers and planners to maintain stability and execute planned policies~\citep{gong2020angular}. The former is the focus of this work.

Proprioceptive state estimators often fuse the measurements from an Inertial Measurement Unit (IMU) with leg odometry. Leg odometry, in particular, uses kinematics and contact data to update the state. Reliable measurements of kinematic and contact information become critical in this case. Not all legged robots are equipped with dedicated contact sensors or springs to detect contact~\citep{bledt2018cheetah,katz2019mini}. The addition of dedicated contact sensors is non-trivial and often leads to hardware redesign. 

In this paper, we develop a deep learning-based contact estimator that does not require dedicated sensors and instead uses joint encoders, kinematics, and IMU data. We create contact data sets using an MIT Mini Cheetah robot
\citep{katz2019mini} on eight different terrains. We further deploy the contact estimator along with a contact-aided invariant extended Kalman filter (InEKF)~\citep{hartley2020contact} and show that the resulting odometry trajectory is comparable to a state-of-the-art visual SLAM algorithm (used as a proxy for ground truth). The contributions of this work are as follows. 1) Open-source contact data sets recorded using an MIT Mini Cheetah; 2) A lightweight learning-based contact estimator that mitigates the need for physical contact sensors for state estimation tasks; 3) A quadruped version of the contact-aided invariant EKF compatible with Lightweight Communications and Marshalling (LCM) \citep{huang2010lcm,moore2009lightweight} interface. The data sets and software are available for download~\footnote{\url{https://github.com/UMich-CURLY/deep-contact-estimator} \\ \url{https://github.com/UMich-CURLY/cheetah_inekf_realtime}}; 4) Experimental results for contact-aided state estimation using an MIT Mini Cheetah on 8 different terrains. A video demonstrating the proposed method can be found at: \url{https://youtu.be/oVbP-Y8xT_E}.

\begin{figure}[t]
    \centering
    \subfloat{
    \includegraphics[width=0.54\columnwidth,trim=2.25cm -1.5cm 1.85cm 1cm,clip]{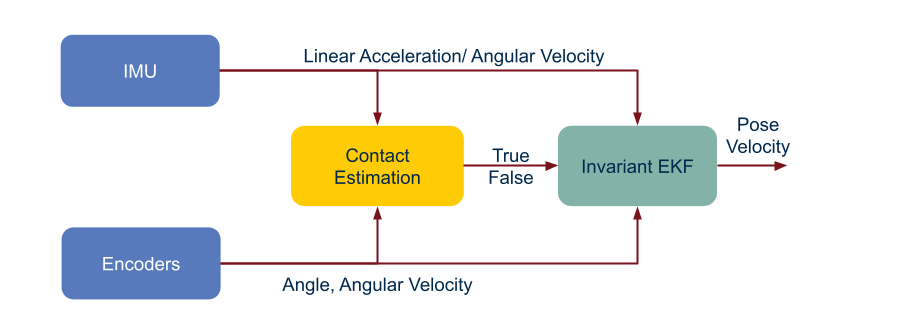}
    }
    \hspace{-0.75cm}\subfloat{
    \includegraphics[width=0.48\columnwidth]{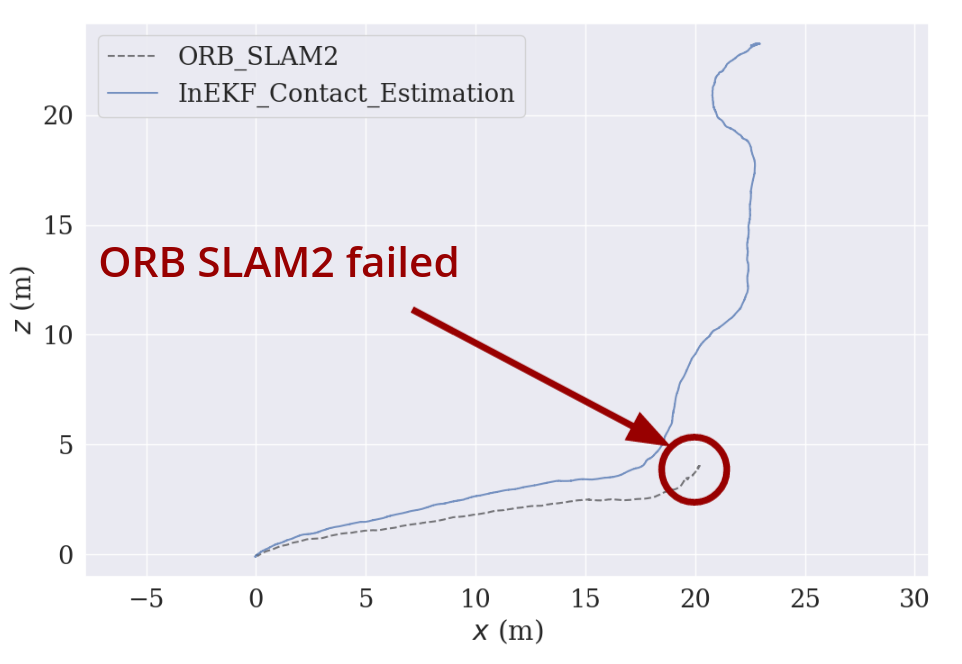}
    }
    \caption{Left: Proposed state estimator. The deep contact estimation network (Yellow) takes joint encoders, IMU, and kinematics data as input and classifies the contact state for the quadruped robot. For state estimation, the estimated contacts, the IMU, and joint encoders data are fused inside a contact-aided invariant extended Kalman filter. Right: The trajectories of the Invariant EKF and ORB SLAM2 in an asphalt and forest data set. The proposed state estimator can serve as reliable odometry when visual SLAM systems fail.}
    \label{fig:frontpage_fig}
    \squeezeup%
\end{figure}

\section{Related Work}
\label{sec:related_work}
Model-based approaches segment the touchdown event of robot legs or prosthetic legs by thresholding on the estimated Ground Reaction Force (GRF) from the general equation of motion~\citep{focchi2013local,fakoorian2016ground,fink2020proprioceptive}. Although this method can detect touchdown events, the estimated GRF is often noisy and unreliable, especially for lightweight robots. \citet{de2006collision,haddadin2008collision} proposed a Generalized Momentum (GM) method for detecting contact events on robot manipulators. This GM-based method is, in fact, a filtered version of the work of~\citet{focchi2013local}. Although GM-based methods mitigate the noise problem in GRF estimation, an empirical threshold on the cut-off frequency is still required. 

\citet{hwangbo2016probabilistic} introduces a probabilistic representation of the contact state and uses a Hidden Markov Model (HMM) to fuse the dynamics and kinematics for contact estimation. They adopt a Monte-Carlo sampling algorithm to compute the transition model and verify the method against GM-based methods. \citet{jenelten2019dynamic} expands the HMM method and focuses on slippage detection. They demonstrate ANYmal~\citep{hutter2016anymal}, a quadruped robot, walking stably on slippery ground. The above two methods aim to detect contact as early as possible for the controller to maintain stability; however, we aim to detect contact intervals for state estimation on various terrains. 

\citet{bledt2018contact} leverages both the GM-based methods and the probabilistic representation of contact states. They use a Kalman filter to fuse the gait phase scheduler information from the controller with the GRF estimated using the GM method and demonstrate that estimates contacts can help the controller reduce the bouncing event upon touchdown. However, this method assumes the leg phase to be periodic as it uses the gait scheduler information in the prediction step of the Kalman filter. It could experience a loss in performance when the phase is heavily violated as the robot interacts with uneven terrains. 

Data-driven approaches take advantage of the rapid development of recent machine learning techniques. \citet{camurri2017probabilistic} uses logistic regression to learn the GRF threshold for contact detection. This work compares against heuristic-based thresholding on GRF using a base state estimator. The result shows that the logistic regression classifier can double the performance of the state estimator. However, compared to deep learning methods, the performance of a logistic regression classifier gets saturated as the number of data increases~\citep{lecun2015deep,Goodfellow-et-al-2016}.
This method requires a specific training procedure for different gait, loading conditions of the robot, and individual terrain properties.  

\citet{rotella2018unsupervised} uses a fuzzy C-means clustering for the probability of contacts in all six end-effector degrees of freedom. They integrate the contact estimator with a base state estimator and show their approach performs considerably better than implementations that are purely based on measured normal force. However, this method assumes contact wrench sensors and additional IMU are available at each end-effector. Furthermore, this method was only tested in simulation. Its performance on real robots remains unknown. \citet{piperakis2019unsupervised} proposes an unsupervised learning method for humanoid gait phase estimation. The authors employ Gaussian Mixture Models (GMMs) for clustering and show the accuracy by comparing to the ground-truth data and leg odometry. However, this work also assumes the availability of wrench/force sensors at each end-effector, and the clustering result is affected by the gait and data density. 

The above methods either assume the availability of wrench/force sensors or are restricted by the nature of simple regression and are thus unable to generalize to different scenarios. In contrast, our work proposes a multi-modal deep learning-based contact estimator that does not require contact sensors and can generalize well to different gaits and terrain properties. Moreover, as more data becomes available, the network performance can be improved.

\section{Preliminaries}
\label{sec:inekf}

\textbf{State Representation.} We wish to estimate the orientation, velocity, and position of the body (IMU) frame in the world frame at any time stamp $t$, which are represented as $R_t \in \mathrm{SO}(3)$, $v_t \in \mathbb{R}^3$, and $p_t \in \mathbb{R}^3$, respectively. 
We define the state $X_t \in \mathrm{SE}_{L+2}(3)$ as $X_t := (R_t, v_t, p_t, d_{lt})$,
where $d_{lt} \in \mathbb{R}^3$ is the position of a contact foot in the world frame. When a contact event is detected for leg $l$, $d_{lt}$ is calculated via the forward kinematics and the current body state, and then augmented into the state to enforce the no-slip (zero-velocity) condition. If the contact constraint breaks, the associated contact position $d_{lt}$ will be marginalized from the state to allow foot movement. The matrix Lie group, $\mathrm{SE}_{L+2}(3)$ is an extension of $\mathrm{SE}(3)$ and it was introduced by~\citet{barrau2015non}.

When a new contact event is detected, the position of the corresponding foot is computed using $\Bar{d}_t = \Bar{p}_t + \Bar{R}_t h_p(\Tilde{\alpha}_t)$, where $h_p(\Tilde{\alpha}_t)$ is the foot position relative to the body frame computed by the forward kinematics (the bar notation denotes estimated variables). $L$ indicates the number of legs that is currently having contact with the ground (here $L \leq 4$).

\textbf{Process Model.} We assume the IMU measurements, i.e., angular velocity and linear acceleration, are corrupted by white Gaussian noise:
\begin{align}
    \nonumber \Tilde{\omega}_t &= \omega_t + w^g_t, \quad w^g_t \sim \mathcal{GP}(0_{3,1},\Sigma^g\delta(t-t')), \\
    \nonumber \Tilde{a}_t &= a_t + w^a_t, \quad w^a_t \sim \mathcal{GP}(0_{3,1},\Sigma^a\delta(t-t')),
\end{align}
where $\mathcal{GP}$ represents a Gaussian process and $\delta(t-t')$ is the Dirac delta function. To handle small foot slippage, we model the contact velocity noise in the body frame via a white Gaussian noise $w^v_t \sim \mathcal{GP}(0_{3,1},\Sigma^v\delta(t-t'))$~\citep{bloesch2012state}.
The process model of the individual term in the state becomes:
\begin{align}
    \frac{d}{dt}R_t &= R_t(\Tilde{\omega}_t-w^g_t)_\times \nonumber, \ 
    \frac{d}{dt}v_t = R_t\Tilde{a}_t-w^a_t+g \nonumber, \\
    \frac{d}{dt}p_t &= v_t, \nonumber \
    \frac{d}{dt}d_t = R_t h_R(\Tilde{\alpha}_t)(-w^v_t) \nonumber ,
\end{align}
where $(\cdot)_\times$ denotes a $3\times3$ skew symmetric matrix, $g$ is the gravity vector, $\Tilde{\alpha}_t$ is the encoder measurements, and $h_R(\Tilde{\alpha}_t)$ is the orientation of the contact frame in IMU (body) frame calculated by forward kinematics using FROST library~\citep{Hereid2017FROST}. 

\textbf{Measurement Model.} We use forward kinematics from current encoder measurements in a right-invariant model described by~\citet{hartley2020contact}. We assume the measurements from joint encoders are corrupted by white Gaussian noise: $\Tilde{\alpha}_t = \alpha + w^\alpha_t, \quad w^\alpha_t \sim \mathcal{N}(0_{3,1},\Sigma^\alpha\delta(t-t'))$. The assumption that the contact point is fixed in the world frame during a contact event leads to $h_p(\Tilde{\alpha}_t) = R_t^\transpose(d_t-p_t)+J_p(\Tilde{\alpha}_t) w^\alpha_t$, where $J_p(\Tilde{\alpha_t})$ is the analytical Jacobian of the forward kinematics function. 

\section{Contact Estimation}
\label{sec:contact_estimation}
In this section, we will discuss our deep learning approach for contact estimation. We define our contact state for each leg $l \in \{RF, LF, RH, LH\}$ as $C = \begin{bmatrix} c_{RF} & c_{LF} & c_{RH} & c_{LH} \end{bmatrix}$ and $c_l \in \{0,1\}$. Here, $0$ denotes no contact, and $1$ indicates a firm contact with the ground. Depending on the robot's motion, the contact state vector $C$ can have totally $16$ different states. (From all feet are in the air to all feet are in contact with the ground). We formulate it as a classification problem in our deep neural network and aim at estimating the correct contact state $C$ given the input data. In order to follow typical classification pipelines, we map our contact state vector $C$ to $16$ different states $S \in \{0,1,\hdots,15\}$ by treating $C$ as a binary value with $4$ digits and using binary to decimal conversion as the function. For example, contact state $C_i = \begin{bmatrix}
    0 & 1 & 1 & 0
\end{bmatrix}$ is mapped to $S_i = 6$.

\subsection{Preprocessing of the Input Data}
The contact estimation network takes sensor measurements from joint encoders, IMU, and kinematics as input. For a synchronized time $n$, the sensor measurements are concatenated as a $54 \times 1$ array $z_n= 
    \begin{bmatrix}
        q_n & \dot{q}_n & a_n & \omega_n & p_{fn} & v_{fn} 
    \end{bmatrix}$,
where
\begin{align*}
    q_n&= 
    \begin{bmatrix}
        q_{RF1n} & q_{RF2n} & q_{RF3n} & q_{LF1n} & \hdots & q_{LH3n}
    \end{bmatrix}\\
    \dot{q}_n&= 
    \begin{bmatrix}
        \dot{q}_{RF1n} & \dot{q}_{RF2n} & \dot{q}_{RF3n} & \dot{q}_{LF1n} & \hdots & \dot{q}_{LH3n}
    \end{bmatrix}\\
    a_n&=
    \begin{bmatrix}
        a_{xn} & a_{yn} & a_{zn}
    \end{bmatrix}, \
    \omega_n=
    \begin{bmatrix}
        \omega_{xn} & \omega_{yn} & \omega_{zn}
    \end{bmatrix}\\
    p_{fn}&= 
    \begin{bmatrix}
        p_{RFxn} & p_{RFyn} & p_{RFzn} & p_{LFxn} & \hdots & p_{LHzn}
    \end{bmatrix}\\
    v_{fn}&= 
    \begin{bmatrix}
        v_{RFxn} & v_{RFyn} & v_{RFzn} & v_{LFxn} & \hdots & v_{LHzn}
    \end{bmatrix}.
\end{align*}
Here, $q_n^\transpose$ is a $12 \times 1$ vector containing all the joint encoder measurements $(\rad)$ at time $n$, $\dot{q}_n^\transpose$ is a $12 \times 1$ vector with joint angular velocity $(\rad/\sec)$, $a_n$ holds the linear accelerations $(\m/\sec)$ from the IMU in the IMU frame, $\omega_n$ contains the angular velocity $(\rad/\sec)$ in the IMU frame, $p_{fn}^\transpose$ is a $12 \times 1$ vector with foot positions calculated from forward kinematics, and $v_{fn}^\transpose$ is a $12 \times 1$ vector that carries the linear velocities of each foot. It is worth noticing that both $p_{fn}$ and $v_{fn}$ are represented in the robot's hip frame. To infer the relationship between data across the time domain, for data point at time $n$, we create a window with size $w$ and append previous measurements within this window into a 2D array $D_n = 
    \begin{bmatrix}
        z_{n-w}^\transpose & z_{n-w+1}^\transpose & \hdots & z_{n}^\transpose 
    \end{bmatrix}^\transpose$.
$D_n$ is a $w \times 54$ array. (In our case, we use $w = 150$.) Each time, the network takes $D_n$ as input and estimates the contact state $S_n$ as output.

\subsection{Network Architecture}
\begin{figure*}[t]
    \includegraphics[width=0.99\textwidth]{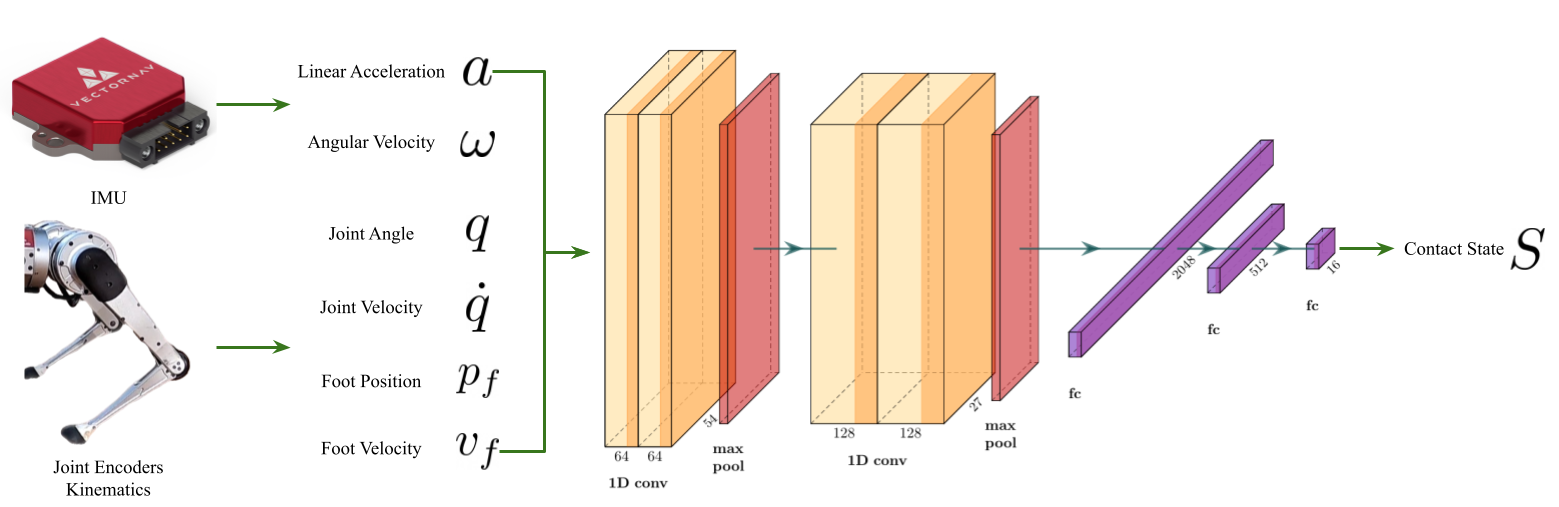}
    \caption{The architecture of the contact estimation network. The network takes in measurements from encoders and an IMU and outputs the contact state of the robot. The structure consists of 2 blocks and 3 fully connected layers. In each block, there are 2 one-dimensional convolution layers and 1 one-dimensional max pooling. The second convolution layer in each block is applied with a dropout mechanism to prevent overfitting.}
    \label{fig:network_structure}
    \squeezeup
\end{figure*}

The contact estimation network consists of 2 blocks of convolutions and 3 fully connected layers, as shown in Figure \ref{fig:network_structure}. Each block contains 2 one-dimensional convolution layers and a one-dimensional max pooling. The convolution layers are designed to extract deep features from the input data. We choose a one-dimensional kernel to increase computational efficiency in terms of memory usage and run time. The kernel moves along the time domain, and the data is padded to preserve dimensions. ReLU is employed as the nonlinear activation function for the convolution layer. The second convolution layer is applied with a dropout mechanism to prevent overfitting. At the end of each block, a one-dimensional max-pooling layer is added to downsample the data. 

The second block is connected to 3 fully connected layers that convert the deep features into the 16 classes we defined earlier. We also employ the dropout mechanisms in the first 2 fully connected layers to prevent overfitting. Finally, we formulate a classification problem using the cross-entropy loss as
    $L(P_i) = -\log{\frac{\exp{(P_i)}}{\sum_j{\exp{(P_j)}}}}$.
$P_j$ is the probability output from the network of state $j$, and $P_i$ is the probability of the ground truth state. The detailed network architecture is listed in the Appendix.

\subsection{Contact Data Set}
\label{subsec:contact_data_set}
We create open-source contact data sets using an MIT Mini Cheetah robot. We record all the proprioceptive sensor measurements from the robot as LCM logs. Recorded measurements include joint encoders data, foot positions and velocities, IMU measurements, and estimated joint torques from an MIT controller~\citep{kim2019highly}. The IMU measurements are received at $1000 \Hz$, while other data are recorded at $500 \Hz$. All data are upsampled to match the IMU frequency after being recorded. Around 1,000,000 data points are collected from various terrains, including asphalt road, concrete, forest, grass, middle pebbles, small pebbles, rock road, and sidewalk. To give the network negative examples of non-contact scenarios, we also record several sequences of the robot walking in the air by holding the robot up and giving the same controller command. (i.e., not having contact with the ground while operating the same gait.)
\begin{table}[t]
    \centering
    \caption{Number of data of each terrain in the contact data sets.}
    \resizebox{0.99\columnwidth}{!}{
    \begin{tabular}{c c c c c c c c c c c}
        \toprule
        \multicolumn{11}{c}{\textbf{Terrain Type}} \\
         overall & air trotting & air pronking & asphalt road & concrete & forest & grass & middle pebble & small pebble & rock road & sidewalk \\
        \midrule
        1,013,441	& 44,386	& 48,972	& 94,615	& 465,144 & 72,144 & 103,392 & 44,442 & 52,669 & 45,819 & 58,115 \\
        
        \bottomrule
     
    \end{tabular}
    }
    \label{tab:data_diff_terrain}
    \squeezeup%
\end{table}

\begin{figure}[t]
    \centering
    \subfloat[Robot Setup for data collection.]{
        \includegraphics[width=0.41\textwidth]{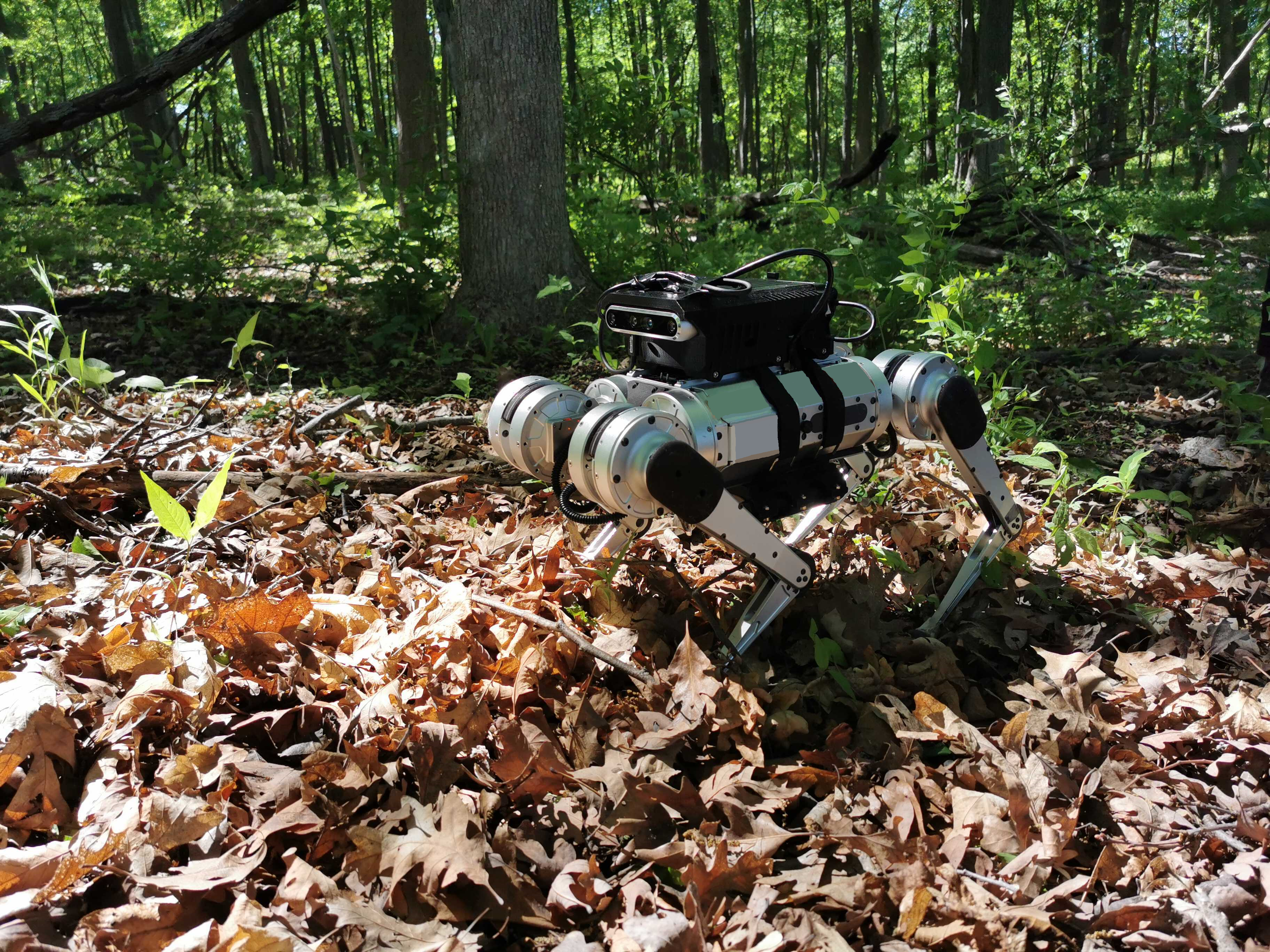}
        \label{fig:lab_setup}}
    \subfloat[Different ground types in the contact data set.]{
        \includegraphics[width=0.55\textwidth,trim={1cm 0.3cm 1cm 1cm},clip]{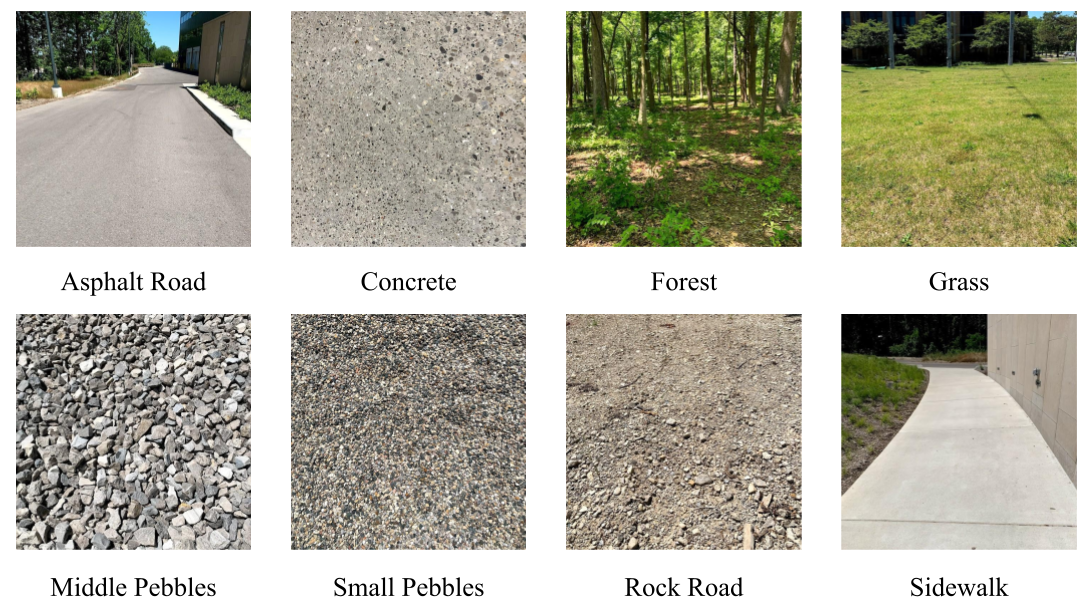}
        \label{fig:groundtype}}
     \caption{(a) Robot configuration for all data sets except grass. Additional RGB-D images are recorded using an Intel D455 camera mounted on the robot. The images are used in ORB SLAM2 to generate ground truth trajectories. (b) Different ground types in the contact data set.}
     \squeezeup
\end{figure}

The grass data sets are collected inside an outdoor facility equipped with a motion capture system, which we use as a proxy for ground truth trajectory. Markers are attached to each foot and the robot body for the motion capture system to record the absolute foot positions and body pose in the world frame. For the rest of the data sets, in addition to the proprioceptive sensors, we also record RGB-D images with an Intel D455 camera mounted on top of the robot. We use these images in a state-of-the-art visual SLAM system, ORB SLAM2~\citep{mur2017orb}, to generate the approximate ground truth trajectories. Figure~\ref{fig:groundtype} shows the picture of each terrain. The concrete data sets are recorded in a lab environment with polished concrete. The forest data sets are collected with random leaves, woods, and plants scattered around the ground. The middle pebbles data sets are recorded on pebbles ranging from $3$ to $7$ cm. The small pebbles data sets are recorded on pebbles ranging from  $0.5$ to $2$ cm. Rock road consists of dirt and rocks of random sizes. The contact data sets contain data from 3 different gaits and different robot loading.

The self-supervised network uses an offline algorithm that automatically generates ground truth labels using robot foot height in the hip frame. As shown in the Appendix, the algorithm applies a low-pass filter to the signal and extracts local minima and maxima using the future and past data points around the current timestamp. Hence, it cannot run online on the robot. The contact events are labeled by connecting the local minima between peaks. Moreover, we observe a bouncing effect on the robot's foot upon touch down after inspecting slow-motion videos of Mini Cheetah's walking patterns. The bouncing results in a sudden change of foot height in the signal. Applying the low-pass filter to the signal enables the algorithm to remove these false positives from the ground truth.

\section{Experimental Results}
\label{sec:experiments}
We train and evaluate the contact network discussed in the previous section. Some sequences are reserved for testing. The rest of the data are split into validation, testing, and training sets with the ratio of $15\%$, $15\%$, and $70\%$, respectively. Table \ref{tab:data_diff_terrain} lists the number of data being used from each terrain. A window size of $150$ is used to allow the network to infer from the time domain. Each data point is normalized along the time domain to prevent scaling issues. We shuffled the training data to ensure generalizability in the model and reduce overfitting. We set the batch size to $30$ and use $10^{-4}$ as the learning rate. The training process took 1.5 hours on an NVIDIA RTX 3090 GPU for 30 epochs. We note that after about seven epochs, the network can converge.

\subsection{Contact Estimation}
We evaluate the performance of the network in two ways. First, 16 class denotes the accuracy in terms of 16 contact states $S$ we defined earlier. This case is a harsher way to evaluate the network since it requires all four legs to be correct simultaneously. We also compute the accuracy of the individual leg by comparing the estimated contact state of each leg with ground truth contacts individually. On average, the trained network achieves $93.88\%$ accuracy in 16 contact states and $97.82\%$ in each leg individually. Detailed accuracy across different terrains is listed in the Appendix. In addition, since false positive in contact estimation is fatal to the state estimator, we compute the False Positive Rate (FPR) and the False Negative Rate (FNR) to obtain insights into the performance. The FPR and the FNR are calculated following the definition $FPR = \frac{FP}{FP+TN}$ and $FNR=\frac{FN}{FN+TP}$, where $FP$ is false positive, $TN$ is true negative, $FN$ is false negative, and $TP$ is true positive. On average, our method holds an FPR of $1.82 \%$ and an FNR of $2.88\%$ on the test sets.

\subsection{Comparison of Contact Detection Methods}
\begin{table}[t]
    \centering
    \caption{Accuracy comparison against baselines. The proposed method achieves the highest accuracy on both sequences. Although the gait cycle method has an accuracy closer to the proposed method, it does not remove false positives when gait cycle is violated.}
    \footnotesize 
    \resizebox{0.99\columnwidth}{!}{
    \begin{tabular}{r r r r r r r c c}
        \toprule
        \textbf{Sequence} & \textbf{Method} & \multicolumn{5}{c}{\textbf{\% Accuracy}}  &
        \multicolumn{1}{c}{\textbf{\% False Positive Rate}} & \multicolumn{1}{c}{\textbf{\% False Negative Rate}} \\
        & &  Leg RF & Leg LF & Leg RH & Leg LH & Leg Avg & Leg Avg & Leg Avg \\
        \midrule
        \multirow{3}{*}{Concrete Short Loop} & GRF Thresholding & 73.43 & 70.02 & 71.69 & 70.04 & 71.30  & 37.07 & 13.24\\
        & Gait Cycle & 85.66 & 84.98 &	84.68 &	85.11 &	85.11 &	22.95 & \textbf{0.00} \\
        & Proposed Method & \textbf{98.34} &	\textbf{97.87} &	\textbf{97.95} &	\textbf{98.56} &	\textbf{98.18} & \textbf{1.45} & 2.51 \\
        \midrule
        \multirow{3}{*}{Grass Test Sequence} & GRF Thresholding & 82.55 & 78.93 & 84.62 &	82.48 &	82.14 & 26.87 & \textbf{0.63} \\
        & Gait Cycle & 92.41 &	92.38 & 91.04 &	90.55 &	91.59 & 10.95 & 3.53 \\
        & Proposed Method & \textbf{98.08} &	\textbf{97.57} &	\textbf{97.73} &	\textbf{97.73} &	\textbf{97.78} & \textbf{2.35} & 1.98 \\
        \midrule
        \multirow{3}{*}{Forest Test Sequence} & GRF Thresholding & 80.99 &	80.09 &	82.75 & 83.24 &	81.77 & 26.54 & 1.84 \\
        & Gait Cycle & 83.03 &	82.56 &	84.44 &	84.28 &	83.58 & 24.71 & \textbf{0.08} \\
        & Proposed Method & \textbf{97.05} &	\textbf{96.62} &	\textbf{97.24} &	\textbf{97.40} &	\textbf{97.08} & \textbf{2.82} & 3.12 \\
        \bottomrule
    \end{tabular}}
    \label{tab:baseline_acc}
    \squeezeup
\end{table}

This experiment compares the accuracy, FPR, and FNR of the proposed method against other contact detection methods using the Mini Cheetah robot. We implement a model-based approach~\citep{focchi2013local,fakoorian2016ground,fink2020proprioceptive} where the estimated ground reaction force is computed via the general equation of motion with a low-pass filter. A fixed threshold is then set to detect the contact events. Because there is no direct access to the motor current on the robot, we use the torque command from the controller to approximate the actual torque on the actuators. We also obtain the gait cycle command from the MIT controller~\citep{kim2019highly} to serve as a contact estimation method.

Table~\ref{tab:baseline_acc} lists the accuracy, FPR, and FNR of the compared methods on three test sequences. We can see that the proposed method achieves the highest accuracy across all sequences and the GRF thresholding has the worst performance. Although the baselines obtain slightly lower FNRs, our method maintains significantly lower FPRs, which is crucial to state estimation tasks.

\subsection{Contact-Aided Invariant Extended Kalman Filter}
We use the contact-aided invariant extended Kalman filter described in Section~\ref{sec:inekf} on concrete test sets. The estimated contacts are converted into LCM messages along with IMU and encoder measurements. The InEKF takes in the LCM messages and estimates the pose of the robot accordingly. We also run the filter using the ground truth contact data to serve as a reference. 

\begin{figure}[t]
    \centering
    \includegraphics[width=0.51\textwidth,trim={2cm 1.4cm 1cm 1cm},clip]{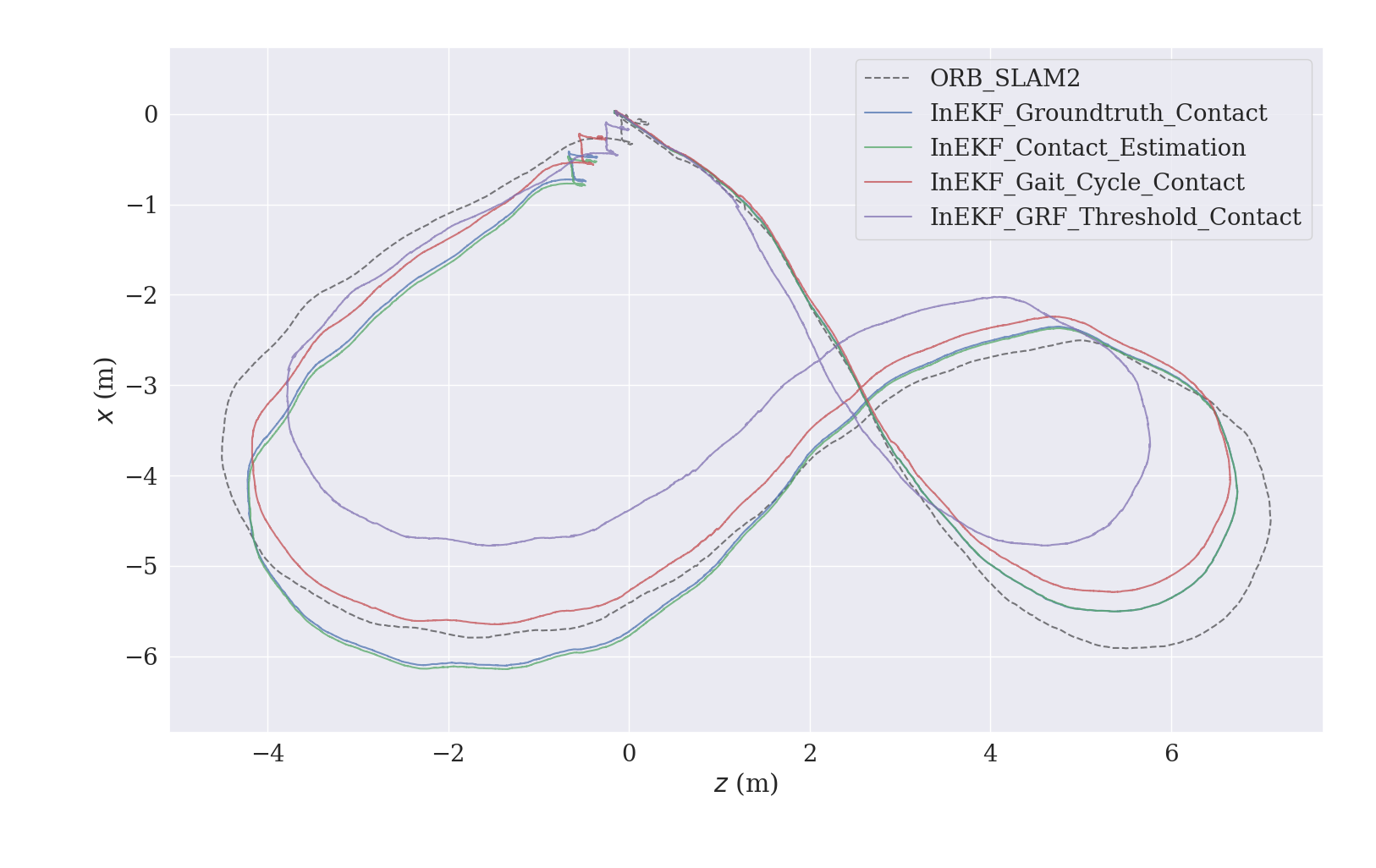}
    \includegraphics[width=0.47\textwidth,trim={1cm 0cm 1cm 1cm},clip]{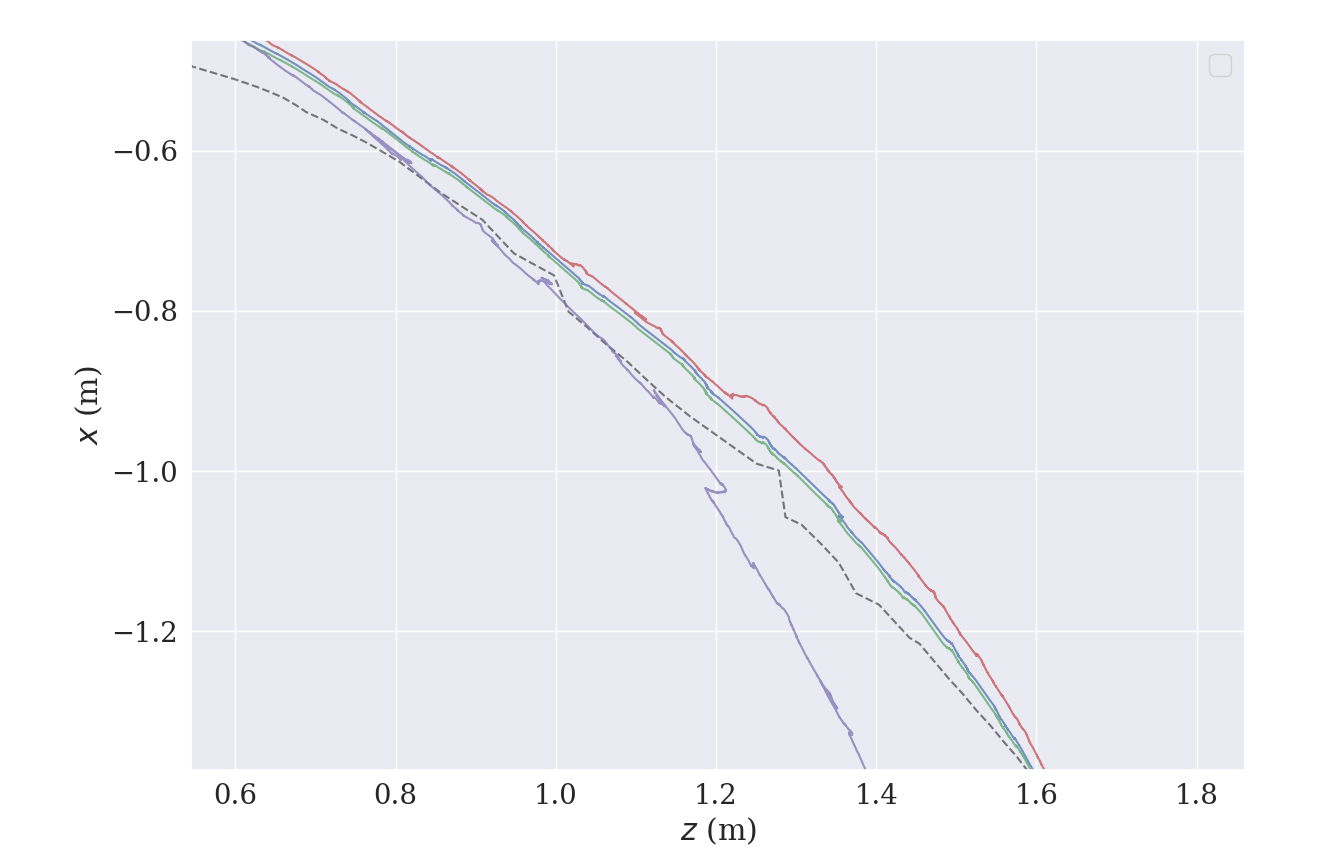}
    \includegraphics[width=0.55\textwidth,trim={1.7cm 0.5cm 1cm 0.5cm},clip]{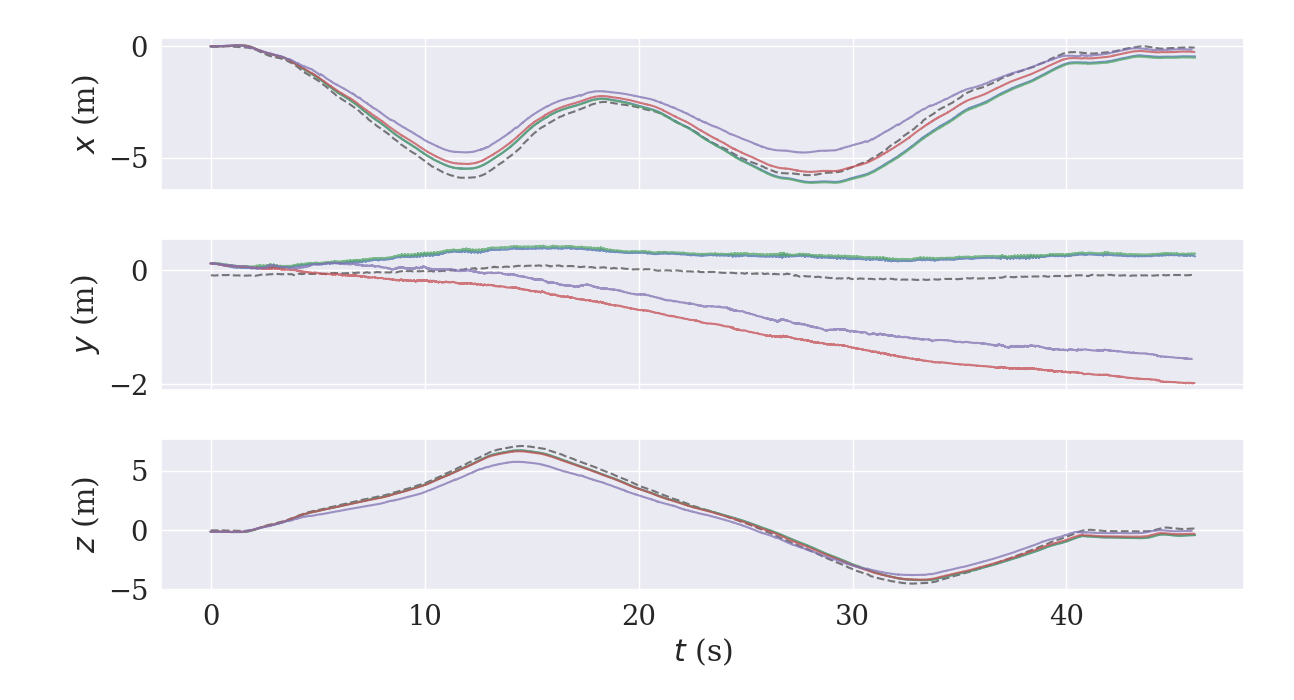}
    \includegraphics[width=0.42\textwidth,trim={0.5cm 0.5cm 0.5cm 0.5cm},clip]{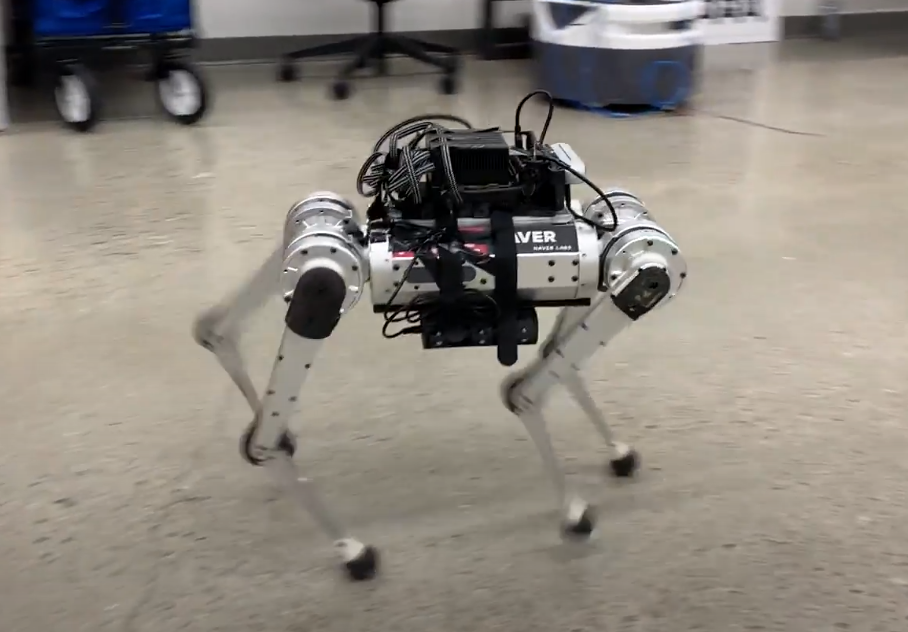}
        
     \caption{Concrete short loop test sequence. Top Left: The bird's-eye view of the trajectories. The estimated trajectory is mapped to the camera frame ($Y$ pointing downward, and $Z$ pointing forward). Using the estimated contacts, the InEKF can reproduce a similar trajectory to that used ground truth contacts. Top Right: Zoomed-in of the bird's-eye view. The baseline contact estimation methods yield zig-zag trajectories, while the proposed method generates a smoother trajectory. Bottom Left: This figure shows that the gait cycle and GRF thresholding methods produce a significant height (Y) drift in the InEKF state estimator. Bottom Right: Robot configuration in this experiment.}
     \label{fig:lab_traj}
\end{figure}

Figure~\ref{fig:lab_traj} shows the trajectories from the InEKF using different contact sources on a concrete loop sequence.  Qualitatively compared to the baseline contact detectors, the resulting trajectory with the proposed contact estimation has smaller drifts from the trajectory with ground truth contacts, especially in the height (Y) axis. Furthermore, compared to the baseline contact estimators, the proposed method also yields a smoother trajectory. %

\subsection{Cassie-Series Robot Experiment}
We also performed an experiment on a Cassie-series bipedal robot, developed by Agility Robotics, to show the applicability of the proposed framework to different robots and training label sources. The robot is equipped with a spring on each leg, which serves as a contact detection sensor. The contact state is defined as $C =          \begin{bmatrix} c_{L} & c_{R} \end{bmatrix}$. The decimal contact states becomes $S \in  \{0,1,\hdots,3\}$. Similar to MiniCheetah, we concatenate the input data as  $z_n= 
    \begin{bmatrix}
        q_n & \dot{q}_n & a_n & \omega_n & p_{fn} & v_{fn} 
    \end{bmatrix}$. Because data from Cassie is recorded at a higher frequency, a window size of 600 is used.

The data is recorded on polished concrete using a controller developed by~\citet{gong2021zero, gong2021one}. Around $1,300,000$ data points are recorded. Two sequences are reserved as test sets, and two are used in training. We obtained two sources of ground truth contacts in this experiment - one from spring deflections; the other from kinematics. Spring contact labels are obtained by thresholding on the GRF computed from the spring deflection. We assume this is a more reliable contact source as it comes from a physical sensor. The kinematics contacts are generated following the same procedure described in Section \ref{subsec:contact_data_set}. We train the network using two ground truth contacts separately and evaluate them against the spring contacts.

\begin{table}[t]
    \centering
    \caption{The accuracy of the network trained with data from a Cassie-series robot. The network is trained with labels generated from spring deflections and labels from kinematics separately. The network trained with kinematics labels maintains low FPR.}
    \footnotesize
    \resizebox{0.9\columnwidth}{!}{\begin{tabular}{r r r r r | r r r | r r r}
        \toprule
        \textbf{Contact Labels} & \multicolumn{4}{c}{\textbf{\% Accuracy}}  & \multicolumn{3}{c}{\textbf{\% False Positive Rate}} & \multicolumn{3}{c}{\textbf{\% False Negative Rate}}\\
         & 4 class & Leg L & Leg R & Leg Avg & Leg L & Leg R & Leg Avg & Leg L & Leg R & Leg Avg\\
        \midrule
        Spring Deflection & 98.03 & 99.82 & 98.19 & 99.01 & 0.13 & 2.18 & 1.18 & 0.23 & 1.36 & 0.78  \\
        Kinematics & 85.14 & 93.05 & 92.09 & 92.57 & 0.01 & 2.42 & 1.24 & 14.56 & 14.55 & 14.55  \\
        \bottomrule
    \end{tabular}}
    \label{tab:acc_cassie}
\end{table}

Table~\ref{tab:acc_cassie} lists the accuracy, the FPR, and the FNR of the two networks. We can see that when trained with ground truth contacts using spring deflection, the network can achieve above $98\%$ accuracy. When trained with the labels generated using kinematics, it can maintain above $92\%$ accuracy on both legs. The results indicate that training using labels generated from the process described in Section~\ref{subsec:contact_data_set} is useful in practice, albeit with some loss of accuracy. Moreover, the network trained with kinematics labels does not increase the FPR, a valuable trait for state estimation tasks. This experiment also shows the generalizability of the proposed framework. We demonstrate that the proposed framework can work on two different robots as long as one collects enough data for training. 

\subsection{Ablation Study and Runtime}

We perform an ablation study on different network architectures to study how the performance changes. Table \ref{tab:acc_diff_network_struc} lists the average performance of different network structure on the test sets. Each "block" consists of 2 convolution layers followed by ReLU activation function and 1 max pooling, as described in Figure \ref{fig:network_structure}. In the table, 2 Blocks is the proposed network detailed in Section \ref{sec:contact_estimation}. 1 Block refers to the network with only the first block and the fully connected layers of the proposed network. As for 4 Blocks, we added two additional blocks with 256 and 512 channels before the fully connected layers. Conv-Pool-Conv-Pool consists of only 2 convolution layers of sizes 64 and 128 and the fully connected layers. Each convolution layer is followed by a ReLU function and one max pooling. From the table, we can see that 2 Blocks has the best performance among all other structures. However, 1 Block maintains a smaller network while showing a slightly lower accuracy.

\begin{table}[t]
    \centering
    \caption{The average performance of different network structures on the test sets. Each "block" consists of 2 convolution layers followed by ReLU activation function and 1 max pooling, as described in Figure~\ref{fig:network_structure}.}
    \footnotesize
    \resizebox{0.99\columnwidth}{!}{\begin{tabular}{r r r r r r r c c }
        \toprule
        \textbf{Network Structure} & \multicolumn{6}{c}{\textbf{\% Accuracy}}  & \multicolumn{1}{c}{\textbf{\% False Positive Rate}} & \multicolumn{1}{c}{\textbf{\% False Negative Rate}} \\
         & 16 class & Leg RF & Leg LF & Leg RH & Leg LH & Leg Avg & Leg Avg & Leg Avg\\
        \midrule
        2 Blocks & \textbf{93.88} & \textbf{97.76} & \textbf{97.66} & \textbf{97.86} & \textbf{98.00} & \textbf{97.82} & \textbf{1.82} & \textbf{2.88} \\
        
        1 Block	& 93.58 & 97.59 & 97.47 & 97.77 & 97.80 & 97.66 & 1.83 & 3.34  \\
        
        4 Blocks & 92.19 & 96.90 & 96.64 & 97.03 & 97.56 & 97.04 & 2.28 & 4.28 \\
        
        Conv-Pool-Conv-Pool & 93.13 & 97.45 & 97.19 & 97.46 & 97.46 & 97.39 & 2.06 & 3.67  \\
        
        \bottomrule
    \end{tabular}}
    \label{tab:acc_diff_network_struc}
\end{table}

The inference speed on an NVIDIA RTX 3090 GPU is approximately 1100 Hz. The inference speed on an NVIDIA Jetson AGX Xavier, which we equipped on the robot, is around 830 Hz after TensorRT optimization. The frequency of the encoder measurements on Mini Cheetah is 500 Hz. This means the proposed contact estimator run at real-time without delay on the robot. The InEKF runs at 2000 Hz on an Intel i7-8750H CPU.

\section{Discussion and Limitations}
\label{sec:discussion}
The developed contact estimator can capture contact events for a quadruped robot without force/contact sensors. However, the contacts are modeled as binary values, and the ``quality'' of contacts is not modeled in our contact estimation network. As a result, a perfect contact is always assumed when a contact event is detected. Furthermore, the covariance matrix of the contact measurements is set heuristically in the InEKF, which along with the uncertainties in the kinematic models, could be the potential sources of estimation error in the InEKF. In the future, we wish to expand the contact detection network to have a covariance estimation of the current contact state. This approach will allow the state estimator to treat contact information as a sensor measurement with an online uncertainty update and potentially decrease the drift in the state estimation. 

We formulated the problem as a classification problem with $16$ different contact states. This model can potentially be further improved by employing a multi-task learning approach~\citep{9392366,evgeniou2004regularized}. This approach allows to independently segment the contact events for each foot.%

\section{Conclusion}
\label{sec:conclusion}
We developed a multi-modal deep learning-based contact estimation method that does not require contact/force sensors and works well with different robot gaits on distinct terrains. We present open-source contact data sets with self-supervised labels of contact events using an MIT Mini Cheetah robot. We show that the trained contact network can achieve $97\%$ accuracy across different terrains. The estimated contacts are employed in a contact-aided invariant state estimator for quadruped robots, and the resulting trajectory is comparable to a modern visual SLAM system.

\clearpage
\section*{Acknowledgment}
Toyota Research Institute provided funds to support this work. Funding for M. Ghaffari was in part provided by NSF Award No. 2118818. This work was also supported by MIT Biomimetic Robotics Lab and NAVER LABS. NVIDIA Corporation provided hardware support for this work.

{\small
\bibliography{bib/strings-full,bib/ieee-full,bib/refs}

\begin{thebibliography}{34}
\providecommand{\natexlab}[1]{#1}
\providecommand{\url}[1]{\texttt{#1}}
\expandafter\ifx\csname urlstyle\endcsname\relax
  \providecommand{\doi}[1]{doi: #1}\else
  \providecommand{\doi}{doi: \begingroup \urlstyle{rm}\Url}\fi

\bibitem[Raibert et~al.(1989)Raibert, Brown~Jr, Chepponis, Koechling, and
  Hodgins]{raibert1989dynamically}
M.~H. Raibert, H.~B. Brown~Jr, M.~Chepponis, J.~Koechling, and J.~K. Hodgins.
\newblock Dynamically stable legged locomotion.
\newblock Technical report, Massachusetts Inst of Tech Cambridge Artificial
  Intelligence Lab, 1989.

\bibitem[Bijelic et~al.(2018)Bijelic, Gruber, and Ritter]{bijelic2018benchmark}
M.~Bijelic, T.~Gruber, and W.~Ritter.
\newblock A benchmark for {LIDAR} sensors in fog: Is detection breaking down?
\newblock In \emph{IEEE Intelligent Vehicles Symposium}, pages 760--767. IEEE,
  2018.

\bibitem[Belter and Nowicki(2019)]{belter2019optimization}
D.~Belter and M.~R. Nowicki.
\newblock Optimization-based legged odometry and sensor fusion for legged robot
  continuous localization.
\newblock \emph{Robotics and Autonomous Systems}, 111:\penalty0 110--124, 2019.

\bibitem[{Gan} et~al.(2020){Gan}, {Zhang}, {Grizzle}, {Eustice}, and
  {Ghaffari}]{gan2019bayesian}
L.~{Gan}, R.~{Zhang}, J.~W. {Grizzle}, R.~M. {Eustice}, and M.~{Ghaffari}.
\newblock Bayesian spatial kernel smoothing for scalable dense semantic
  mapping.
\newblock \emph{IEEE Robotics and Automation Letters}, 5\penalty0 (2):\penalty0
  790--797, April 2020.
\newblock ISSN 2377-3774.

\bibitem[Wisth et~al.(2020)Wisth, Camurri, and Fallon]{wisth2020preintegrated}
D.~Wisth, M.~Camurri, and M.~Fallon.
\newblock Preintegrated velocity bias estimation to overcome contact
  nonlinearities in legged robot odometry.
\newblock In \emph{Proceedings of the {IEEE} International Conference on
  Robotics and Automation}, pages 392--398. IEEE, 2020.

\bibitem[Gong and Grizzle(2020)]{gong2020angular}
Y.~Gong and J.~Grizzle.
\newblock Angular momentum about the contact point for control of bipedal
  locomotion: Validation in a lip-based controller.
\newblock \emph{arXiv preprint arXiv:2008.10763}, 2020.

\bibitem[Bledt et~al.(2018)Bledt, Powell, Katz, Di~Carlo, Wensing, and
  Kim]{bledt2018cheetah}
G.~Bledt, M.~J. Powell, B.~Katz, J.~Di~Carlo, P.~M. Wensing, and S.~Kim.
\newblock {MIT Cheetah} 3: Design and control of a robust, dynamic quadruped
  robot.
\newblock In \emph{Proceedings of the {IEEE}/{RSJ} International Conference on
  Intelligent Robots and Systems}, pages 2245--2252. IEEE, 2018.

\bibitem[Katz et~al.(2019)Katz, Di~Carlo, and Kim]{katz2019mini}
B.~Katz, J.~Di~Carlo, and S.~Kim.
\newblock Mini cheetah: A platform for pushing the limits of dynamic quadruped
  control.
\newblock In \emph{Proceedings of the {IEEE} International Conference on
  Robotics and Automation}, pages 6295--6301. IEEE, 2019.

\bibitem[Hartley et~al.(2020)Hartley, Ghaffari, Eustice, and
  Grizzle]{hartley2020contact}
R.~Hartley, M.~Ghaffari, R.~M. Eustice, and J.~W. Grizzle.
\newblock Contact-aided invariant extended kalman filtering for robot state
  estimation.
\newblock \emph{International Journal of Robotics Research}, 39\penalty0
  (4):\penalty0 402--430, 2020.

\bibitem[Huang et~al.(2010)Huang, Olson, and Moore]{huang2010lcm}
A.~S. Huang, E.~Olson, and D.~C. Moore.
\newblock Mlc: Lightweight communications and marshalling.
\newblock In \emph{Proceedings of the {IEEE}/{RSJ} International Conference on
  Intelligent Robots and Systems}, pages 4057--4062. IEEE, 2010.

\bibitem[Moore et~al.(2009)Moore, Olson, and Huang]{moore2009lightweight}
D.~Moore, E.~Olson, and A.~Huang.
\newblock Lightweight communications and marshalling for low-latency
  interprocess communication.
\newblock 2009.

\bibitem[Focchi et~al.(2013)Focchi, Barasuol, Havoutis, Buchli, Semini, and
  Caldwell]{focchi2013local}
M.~Focchi, V.~Barasuol, I.~Havoutis, J.~Buchli, C.~Semini, and D.~G. Caldwell.
\newblock Local reflex generation for obstacle negotiation in quadrupedal
  locomotion.
\newblock In \emph{Nature-Inspired Mobile Robotics}, pages 443--450. World
  Scientific, 2013.

\bibitem[Fakoorian et~al.(2016)Fakoorian, Simon, Richter, and
  Azimi]{fakoorian2016ground}
S.~A. Fakoorian, D.~Simon, H.~Richter, and V.~Azimi.
\newblock Ground reaction force estimation in prosthetic legs with an extended
  kalman filter.
\newblock In \emph{2016 Annual IEEE Systems Conference (SysCon)}, pages 1--6.
  IEEE, 2016.

\bibitem[Fink and Semini(2020)]{fink2020proprioceptive}
G.~Fink and C.~Semini.
\newblock Proprioceptive sensor fusion for quadruped robot state estimation.
\newblock In \emph{Proceedings of the {IEEE}/{RSJ} International Conference on
  Intelligent Robots and Systems}, pages 10914--10920. IEEE, 2020.

\bibitem[De~Luca et~al.(2006)De~Luca, Albu-Schaffer, Haddadin, and
  Hirzinger]{de2006collision}
A.~De~Luca, A.~Albu-Schaffer, S.~Haddadin, and G.~Hirzinger.
\newblock Collision detection and safe reaction with the dlr-iii lightweight
  manipulator arm.
\newblock In \emph{Proceedings of the {IEEE}/{RSJ} International Conference on
  Intelligent Robots and Systems}, pages 1623--1630. IEEE, 2006.

\bibitem[Haddadin et~al.(2008)Haddadin, Albu-Schaffer, De~Luca, and
  Hirzinger]{haddadin2008collision}
S.~Haddadin, A.~Albu-Schaffer, A.~De~Luca, and G.~Hirzinger.
\newblock Collision detection and reaction: A contribution to safe physical
  human-robot interaction.
\newblock In \emph{Proceedings of the {IEEE}/{RSJ} International Conference on
  Intelligent Robots and Systems}, pages 3356--3363. IEEE, 2008.

\bibitem[Hwangbo et~al.(2016)Hwangbo, Bellicoso, Fankhauser, and
  Hutter]{hwangbo2016probabilistic}
J.~Hwangbo, C.~D. Bellicoso, P.~Fankhauser, and M.~Hutter.
\newblock Probabilistic foot contact estimation by fusing information from
  dynamics and differential/forward kinematics.
\newblock In \emph{Proceedings of the {IEEE}/{RSJ} International Conference on
  Intelligent Robots and Systems}, pages 3872--3878. IEEE, 2016.

\bibitem[Jenelten et~al.(2019)Jenelten, Hwangbo, Tresoldi, Bellicoso, and
  Hutter]{jenelten2019dynamic}
F.~Jenelten, J.~Hwangbo, F.~Tresoldi, C.~D. Bellicoso, and M.~Hutter.
\newblock Dynamic locomotion on slippery ground.
\newblock \emph{IEEE Robotics and Automation Letters}, 4\penalty0 (4):\penalty0
  4170--4176, 2019.

\bibitem[Hutter et~al.(2016)Hutter, Gehring, Jud, Lauber, Bellicoso, Tsounis,
  Hwangbo, Bodie, Fankhauser, Bloesch, et~al.]{hutter2016anymal}
M.~Hutter, C.~Gehring, D.~Jud, A.~Lauber, C.~D. Bellicoso, V.~Tsounis,
  J.~Hwangbo, K.~Bodie, P.~Fankhauser, M.~Bloesch, et~al.
\newblock Anymal-a highly mobile and dynamic quadrupedal robot.
\newblock In \emph{Proceedings of the {IEEE}/{RSJ} International Conference on
  Intelligent Robots and Systems}, pages 38--44. IEEE, 2016.

\bibitem[Bledt et~al.(2018)Bledt, Wensing, Ingersoll, and
  Kim]{bledt2018contact}
G.~Bledt, P.~M. Wensing, S.~Ingersoll, and S.~Kim.
\newblock Contact model fusion for event-based locomotion in unstructured
  terrains.
\newblock In \emph{Proceedings of the {IEEE} International Conference on
  Robotics and Automation}, pages 4399--4406. IEEE, 2018.

\bibitem[Camurri et~al.(2017)Camurri, Fallon, Bazeille, Radulescu, Barasuol,
  Caldwell, and Semini]{camurri2017probabilistic}
M.~Camurri, M.~Fallon, S.~Bazeille, A.~Radulescu, V.~Barasuol, D.~G. Caldwell,
  and C.~Semini.
\newblock Probabilistic contact estimation and impact detection for state
  estimation of quadruped robots.
\newblock \emph{IEEE Robotics and Automation Letters}, 2\penalty0 (2):\penalty0
  1023--1030, 2017.

\bibitem[LeCun et~al.(2015)LeCun, Bengio, and Hinton]{lecun2015deep}
Y.~LeCun, Y.~Bengio, and G.~Hinton.
\newblock Deep learning.
\newblock \emph{Nature}, 521\penalty0 (7553):\penalty0 436--444, 2015.

\bibitem[Goodfellow et~al.(2016)Goodfellow, Bengio, and
  Courville]{Goodfellow-et-al-2016}
I.~Goodfellow, Y.~Bengio, and A.~Courville.
\newblock \emph{Deep Learning}.
\newblock MIT Press, 2016.
\newblock \url{http://www.deeplearningbook.org}.

\bibitem[Rotella et~al.(2018)Rotella, Schaal, and
  Righetti]{rotella2018unsupervised}
N.~Rotella, S.~Schaal, and L.~Righetti.
\newblock Unsupervised contact learning for humanoid estimation and control.
\newblock In \emph{Proceedings of the {IEEE} International Conference on
  Robotics and Automation}, pages 411--417. IEEE, 2018.

\bibitem[Piperakis et~al.(2019)Piperakis, Timotheatos, and
  Trahanias]{piperakis2019unsupervised}
S.~Piperakis, S.~Timotheatos, and P.~Trahanias.
\newblock Unsupervised gait phase estimation for humanoid robot walking.
\newblock In \emph{Proceedings of the {IEEE} International Conference on
  Robotics and Automation}, pages 270--276. IEEE, 2019.

\bibitem[Barrau(2015)]{barrau2015non}
A.~Barrau.
\newblock \emph{Non-linear state error based extended {Kalman} filters with
  applications to navigation}.
\newblock PhD thesis, Mines Paristech, 2015.

\bibitem[Bloesch et~al.(2012)Bloesch, Hutter, Hoepflinger, Leutenegger,
  Gehring, Remy, and Siegwart]{bloesch2012state}
M.~Bloesch, M.~Hutter, M.~A. Hoepflinger, S.~Leutenegger, C.~Gehring, C.~Remy,
  and R.~Siegwart.
\newblock State estimation for legged robots: Consistent fusion of leg
  kinematics and {IMU}.
\newblock In \emph{Proceedings of the Robotics: Science and Systems
  Conference}, 2012.

\bibitem[Hereid and Ames(2017)]{Hereid2017FROST}
A.~Hereid and A.~D. Ames.
\newblock {FROST}: Fast robot optimization and simulation toolkit.
\newblock In \emph{Proceedings of the {IEEE}/{RSJ} International Conference on
  Intelligent Robots and Systems}, Vancouver, BC, Canada, 2017. IEEE.

\bibitem[Kim et~al.(2019)Kim, Di~Carlo, Katz, Bledt, and Kim]{kim2019highly}
D.~Kim, J.~Di~Carlo, B.~Katz, G.~Bledt, and S.~Kim.
\newblock Highly dynamic quadruped locomotion via whole-body impulse control
  and model predictive control.
\newblock \emph{arXiv preprint arXiv:1909.06586}, 2019.

\bibitem[Mur-Artal and Tard{\'o}s(2017)]{mur2017orb}
R.~Mur-Artal and J.~D. Tard{\'o}s.
\newblock {ORB-SLAM2}: An open-source slam system for monocular, stereo, and
  {RGB-D} cameras.
\newblock \emph{{IEEE} Transactions on Robotics}, 33\penalty0 (5):\penalty0
  1255--1262, 2017.

\bibitem[Gong and Grizzle(2021{\natexlab{a}})]{gong2021zero}
Y.~Gong and J.~Grizzle.
\newblock Zero dynamics, pendulum models, and angular momentum in feedback
  control of bipedal locomotion.
\newblock \emph{arXiv preprint arXiv:2105.08170}, 2021{\natexlab{a}}.

\bibitem[Gong and Grizzle(2021{\natexlab{b}})]{gong2021one}
Y.~Gong and J.~Grizzle.
\newblock One-step ahead prediction of angular momentum about the contact point
  for control of bipedal locomotion: Validation in a lip-inspired controller.
\newblock In \emph{2021 IEEE International Conference on Robotics and
  Automation (ICRA)}, pages 2832--2838. IEEE, 2021{\natexlab{b}}.

\bibitem[Zhang and Yang(2021)]{9392366}
Y.~Zhang and Q.~Yang.
\newblock A survey on multi-task learning.
\newblock \emph{IEEE Transactions on Knowledge and Data Engineering}, pages
  1--1, 2021.
\newblock \doi{10.1109/TKDE.2021.3070203}.

\bibitem[Evgeniou and Pontil(2004)]{evgeniou2004regularized}
T.~Evgeniou and M.~Pontil.
\newblock Regularized multi--task learning.
\newblock In \emph{Proceedings of the tenth ACM SIGKDD international conference
  on Knowledge discovery and data mining}, pages 109--117, 2004.

\end{thebibliography}
}

\newpage
\section{Appendix}
\subsection{Network Structure}
Table \ref{tab:network_structure} lists the detailed structure of the proposed contact estimator.
\begin{table}[h!]
    \centering
    \caption{The proposed contact estimation network architecture. Conv1D stands for 1-dimensional convolution layer, Pool1D means 1-dimensional max pooling, and FC is the fully connected layer.}
    \tiny 
    \begin{tabular}{c c c c c c}
        \toprule 
        & Operator & Stride & Filters & Size & Output Shape \\
        \midrule 
        & Input & & & & $N \times 54 \times w$\\
        \midrule  
        \parbox[t]{3mm}{\multirow{3}{*}{\rotatebox[origin=c]{90}{\textbf{Block 1}}}}  
         & Conv1D & 1 & 64 & 3 & $N \times 64 \times w$ \\
         & Conv1D & 1 & 64 & 3 & $N \times 64 \times w$ \\
         & Pool1D & 2 & 64 & 2 & $N \times 64 \times w/2 $ \\
         \midrule
         \parbox[t]{3mm}{\multirow{3}{*}{\rotatebox[origin=c]{90}{\textbf{Block 2}}}}
         & Conv1D & 1 & 128 & 3 & $N \times 128 \times w/2$\\
         & Conv1D & 1 & 128 & 3 & $N \times 128 \times w/2$\\
         & Pool1D & 2 & 128 & 2 & $N \times 128 \times w/4$\\ 
         \midrule
         \parbox[t]{3mm}{\multirow{3}{*}{\rotatebox[origin=c]{90}{\textbf{FC}}}}
         & FC1 & - & - & - & 2048\\
         & FC2 & - & - & - & 512\\ 
         & FC3 & - & - & - & 16\\
         \bottomrule
    \end{tabular}
    \label{tab:network_structure}
    \squeezeup\squeezeup
\end{table}
\subsection{Algorithm for Groundturth Contact Generation}
\begin{figure}[h]
    \centering
    \includegraphics[width=0.99\textwidth]{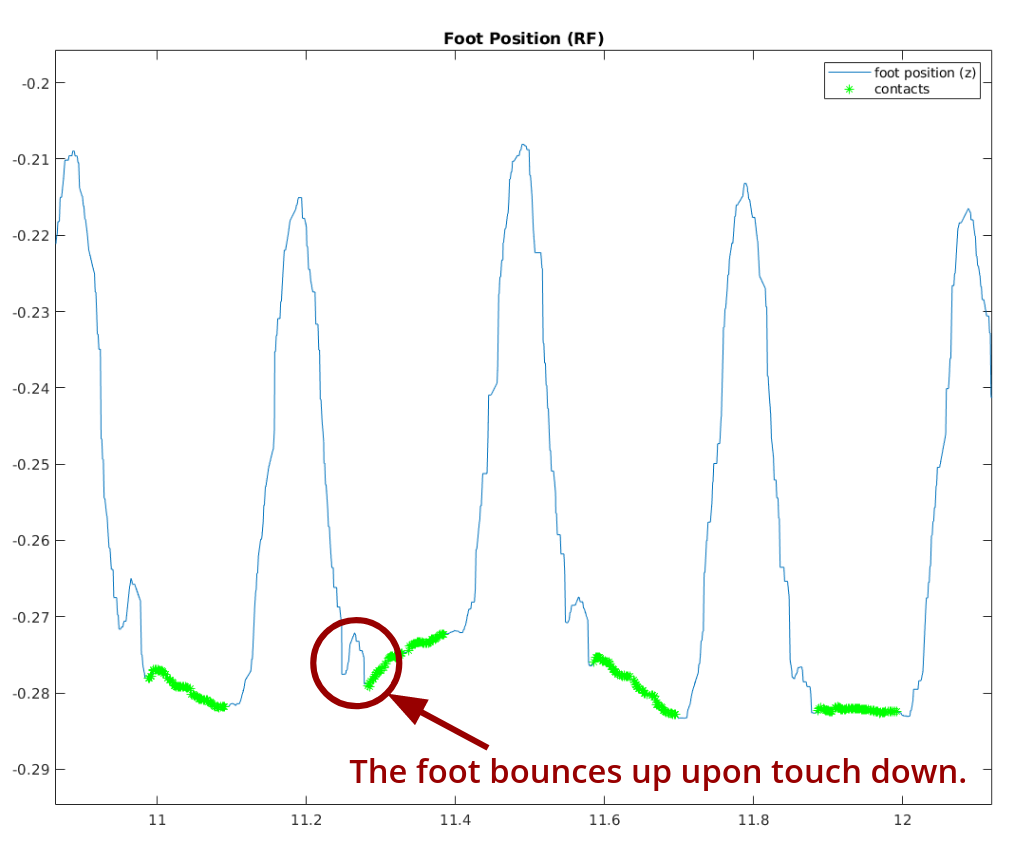}
    \caption{An example of labeled ground truth contacts (in green) overlapped with foot height measurements (in blue) in the hip frame.  As seen in the figure, the initial "jump" in the signal is removed from contact ground truth labels.}
    \label{fig:contact_gt_example}
\end{figure}

The algorithm used to generate the ground truth labels is listed in Algorithm \ref{algo_gt_generation}. The algorithm connects the local minimum between peaks after applying a low-pass filter to the foot height signal. An example of the ground truth contacts overlapping with foot height is shown in Figure \ref{fig:contact_gt_example}. From the plot, we can see the foot bounces up upon touch down, and the algorithm is able to remove it from the ground truth contact list.

\begin{algorithm}
\caption{Ground Truth Label Generation}
\label{algo_gt_generation}
\begin{algorithmic}[1]
\State \texttt{contacts} = \textbf{False}(\texttt{num\_data}, \texttt{num\_legs})
\For {\texttt{all legs}}\\
    
    \State \# Set different half power frequency based on gait
    \If {\texttt{gait} == \texttt{trot}}
        \State \texttt{half_power_freq} = 0.04 
    \ElsIf{\texttt{gait} == \texttt{(pronking or gallop)}}
        \State \texttt{half_power_freq} = 0.08
    \EndIf\\
    
    \State \texttt{foot_height} = \textbf{low_pass_filter}(\texttt{foot_height}, \texttt{half_power_freq})\\
    
    \State \# Extract local max and local min
    \State \texttt{local\_max} = \textbf{is\_local\_max}(\texttt{foot\_height}) 
    \State \texttt{local\_min} = \textbf{is\_local\_min}(\texttt{foot\_height}) \\
    
    \State \# Extract indices from local max and min
    \State \texttt{max_idx} = \textbf{find_idx}(\texttt{local_max})
    \State \texttt{min_idx} = \textbf{find_idx}(\texttt{local_min})\\
    
    \State \texttt{i} = $0$
    \State \texttt{j} = $0$
    \While{\texttt{i} < \texttt{num\_min} \& \texttt{j} < \texttt{num\_max}}
        \State \texttt{contact_start} = \texttt{min_idx}[\texttt{i}]
        \State \texttt{next_peak} = \texttt{max_idx}[\texttt{j}]\\
        
        \State \# Connect all the local minimum before next peak
        \State \texttt{count} = $0$
        \While{\texttt{i} < \texttt{num\_min} \& \texttt{min_idx}[\texttt{i}] < \texttt{next_peak}}
            \State \texttt{contact_end} = \texttt{min_idx}[\texttt{i}]
            \State \texttt{i} = \texttt{i} + $1$
            \State \texttt{count} = \texttt{count} + $1$
        \EndWhile \\
        
        \State \# If only one local minimum is found between two peaks, 
        \State we set a conservative amount of data beforehand as contact.
        \If {\texttt{count} == $1$}
           \State \texttt{contact_start} = \texttt{contact_end} - $30$;
           \If {\texttt{contact_start}<$1$}
               \texttt{contact_start}=$1$;
           \EndIf
        \EndIf
        
        \State \texttt{contacts}[\texttt{contact_start}:\texttt{contact_end}, \texttt{l}] = \textbf{True}
        \State \texttt{j} = \texttt{j} + $1$
    \EndWhile

\EndFor
\end{algorithmic}
\end{algorithm}

\subsection{Contact Estimation Results}
Table \ref{tab:acc_diff_terrain} lists the accuracy of the trained network on different terrains on the test sets. We evaluate the performance of the network in two ways. First, 16 class denotes the accuracy in terms of 16 contact states $S$ we defined earlier. This case is a harsher way to evaluate the network since it requires all four legs to be correct simultaneously. Second, Leg RF to Leg LH list the accuracy of the individual leg by comparing the estimated contact state of each leg with ground truth contacts individually. We can see that, on average, the trained network achieves $93.88\%$ of accuracy in terms of 16 contact states and $97.82\%$ in each leg individually. Across different terrains, the network keeps an accuracy above $96\%$ for each leg. Furthermore, the air trotting and pronking data sets show that when the robot is walking with a similar kinematic pattern in the air, the network can tell there exist no contact event.

\begin{table}[t]
    \centering
    \caption{Accuracy and precision of the proposed network on the test sets. 16 class denotes the accuracy and precision in terms of 16 contact states $S$. Leg RF to Leg LH list the accuracy and precision of individual leg by comparing the correctness of each leg individually.}
        {
        \resizebox{0.96\columnwidth}{!}{
        \begin{tabular}{r r r r r r r c c}
        \toprule
        \textbf{Terrain} & \multicolumn{6}{c}{\textbf{\% Accuracy}} & \multicolumn{1}{c}{\textbf{\% False Positive Rate}} & \multicolumn{1}{c}{\textbf{\% False Negative Rate}}\\
         & 16 class & Leg RF & Leg LF & Leg RH & Leg LH & Leg Avg & Leg Avg & Leg Avg\\
        \midrule
        overall	& 93.88 & 97.76 & 97.66 & 97.86 & 98.00 & 97.82 & 1.82 & 2.88 \\
        
        air trotting 	& 100.00 & 100.00 & 100.00 & 100.00 & 100.00 & 100.00 & 0.00 & N/A \\
        
        air pronking & 100.00 & 100.00 & 100.00 & 100.00 & 100.00 & 100.00 & 0.00 & N/A\\
        
        asphalt road & 94.82 & 98.01 & 98.49 & 98.41 & 98.59 & 98.37 & 1.41 & 2.02 \\
        
        concrete & 93.09 & 97.30 & 97.32 & 97.62 & 97.57 & 97.45 & 2.07 & 3.25\\
        
        forest & 91.26 & 96.88 & 97.47 & 97.51 & 97.38& 97.31 & 2.22 & 3.58\\
        
        grass &  93.60 & 98.13 & 97.57 & 97.73 & 97.80 & 97.81 & 2.33 & 1.93\\
        
        middle pebble &  93.94 & 98.08 &  97.62 & 97.73 & 98.22 & 97.91 & 1.88 & 2,47\\
        
        small pebble &  92.62 & 97.82 & 97.11 & 97.67 & 98.00 & 97.65 & 2.55 & 1.95\\
        
        rock road &  95.45 & 98.83 & 98.39 & 98.46 & 98.90 & 98.65 & 1.12 & 1.79\\
        
        sidewalk &  95.16 & 98.20 & 98.02 & 98.17 & 98.41 & 98.20 & 1.51 & 2.33\\ \bottomrule
     
    \end{tabular}}
    \label{tab:acc_diff_terrain}
        }

\end{table}

\subsection{Contact Estimation with Foot Velocity and Ground Reaction Force}
\begin{figure*}[t]
    \centering
    \includegraphics[trim=4cm 1cm 3cm 1cm,clip,width=\textwidth]{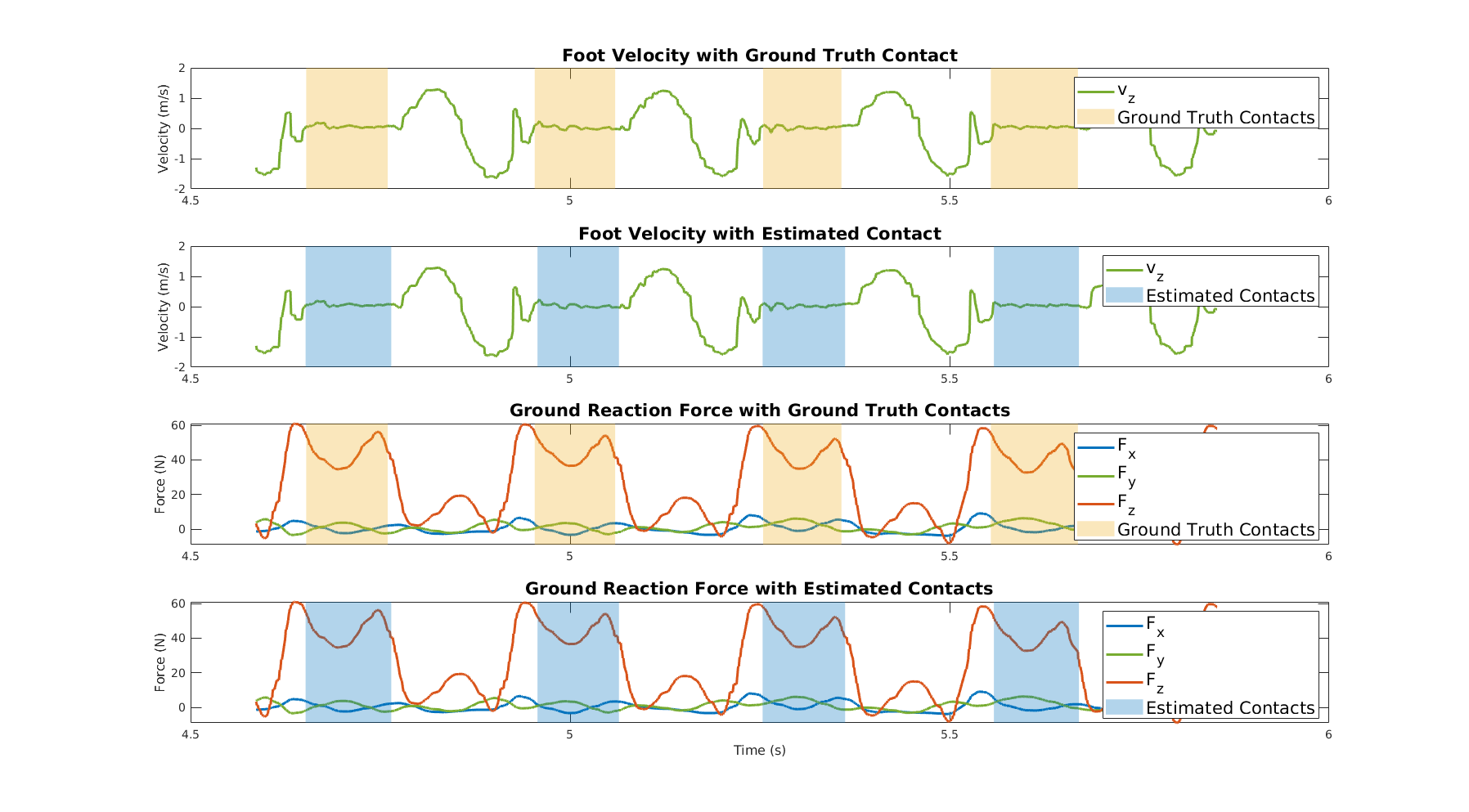}
    \caption{Estimated ground reaction force (After passing through a low-pass filter.) and foot velocity overlapped with estimated contacts and ground truth contacts of one leg in the forest data set. The plots show that the estimated contact phase is consistent with the ground truth contacts. Upon initial contact, the foot bounces up because of the high impact. This plot shows the network can exclude the bouncing phase and correctly estimate contacts where only minimal velocity between the foot and robot body are present.} 
    \label{fig:contact_forest}
\end{figure*}

Figure \ref{fig:contact_forest} shows the estimated contacts overlapped with the estimated ground reaction force and the foot velocity, as well as the ground truth contacts in the forest test set. The yellow sections indicate the ground truth contact phase, and the blue sections are the estimated contact phase. From the plot, we can see that the estimated contacts are consistent with the ground truth contacts. The bottom two plots show the estimated ground reaction force overlapped with contacts, and the top two plots demonstrate the foot velocity versus contacts. It is worth noticing that the foot velocity here is represented in the robot hip frame because we are unable to measure the foot velocity in the world frame directly. The plots show that the estimated contact periods match the periods where only minimal vertical velocity is present between the foot and the robot body. Note that when the foot is first in contact with the ground, the foot bounces up and creates a vertical jump in the foot velocity curve. We conjecture this is because Mini Cheetah's feet are made of rubber balls and are bouncy given the high impact at the initial contact. We wish to have a reliable contact estimation for the state estimator. Thus, even though the estimated ground reaction force is high, the initial contact phase is excluded from the ground truth contacts to avoid violating the no-slip condition. As one can see from the plot, the network can exclude the initial bouncing phase and estimate contact events where the foot velocity is closer to 0.

\begin{figure}[t]
    \centering
    \includegraphics[width=0.85\columnwidth,trim={2cm 1cm 1cm 1cm},clip]{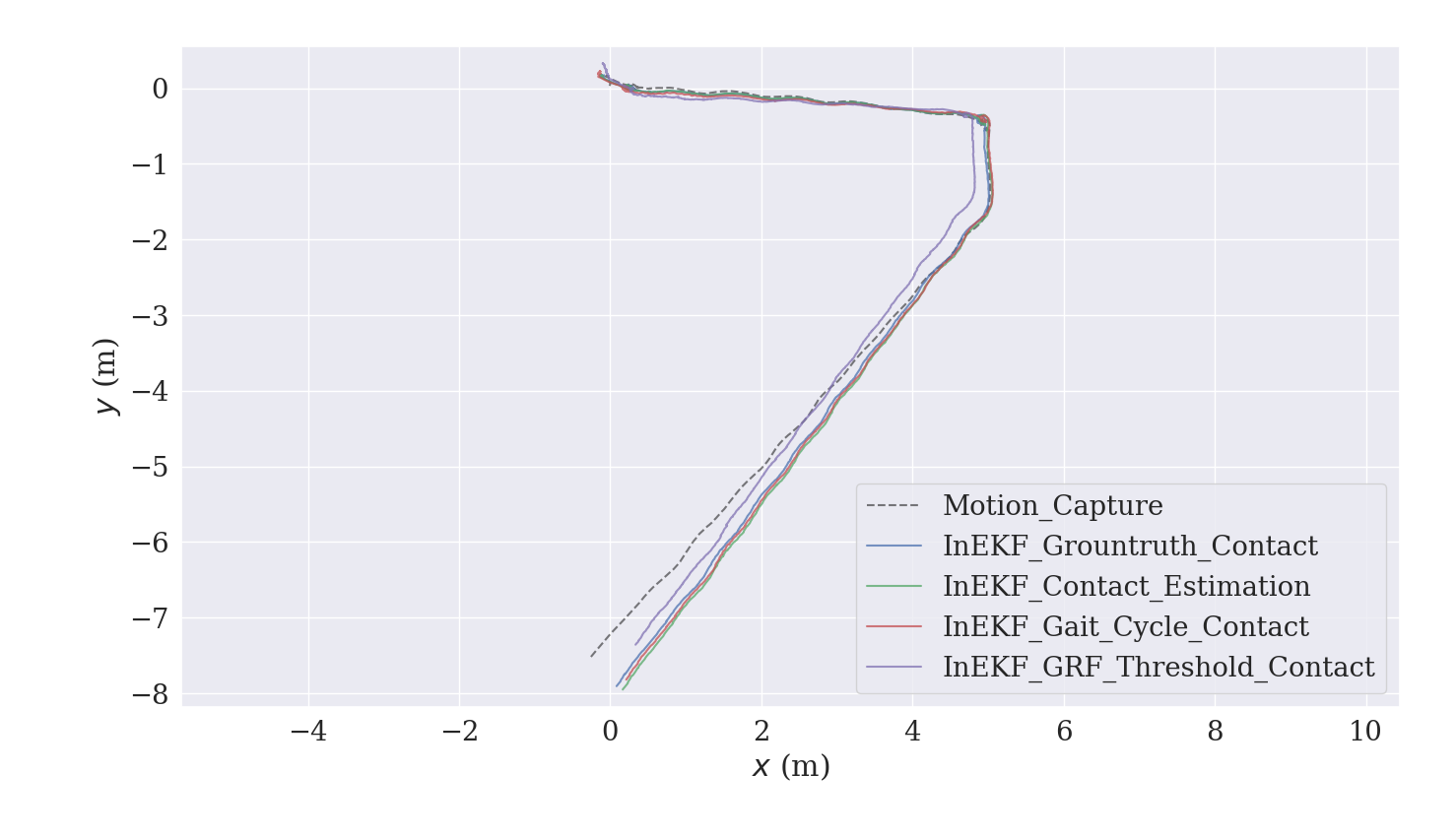}
    \caption{The bird-eye view of the trajectories on the grass test sequence. We can see that the resulting trajectory from GRF thresholding gives a larger drifts in the X-Y plane.}
    \label{fig:mair_baseline_xy}
\end{figure}
\begin{figure}[t]
    \centering
    \includegraphics[width=0.85\columnwidth,trim={2cm 1cm 1cm 1cm},clip]{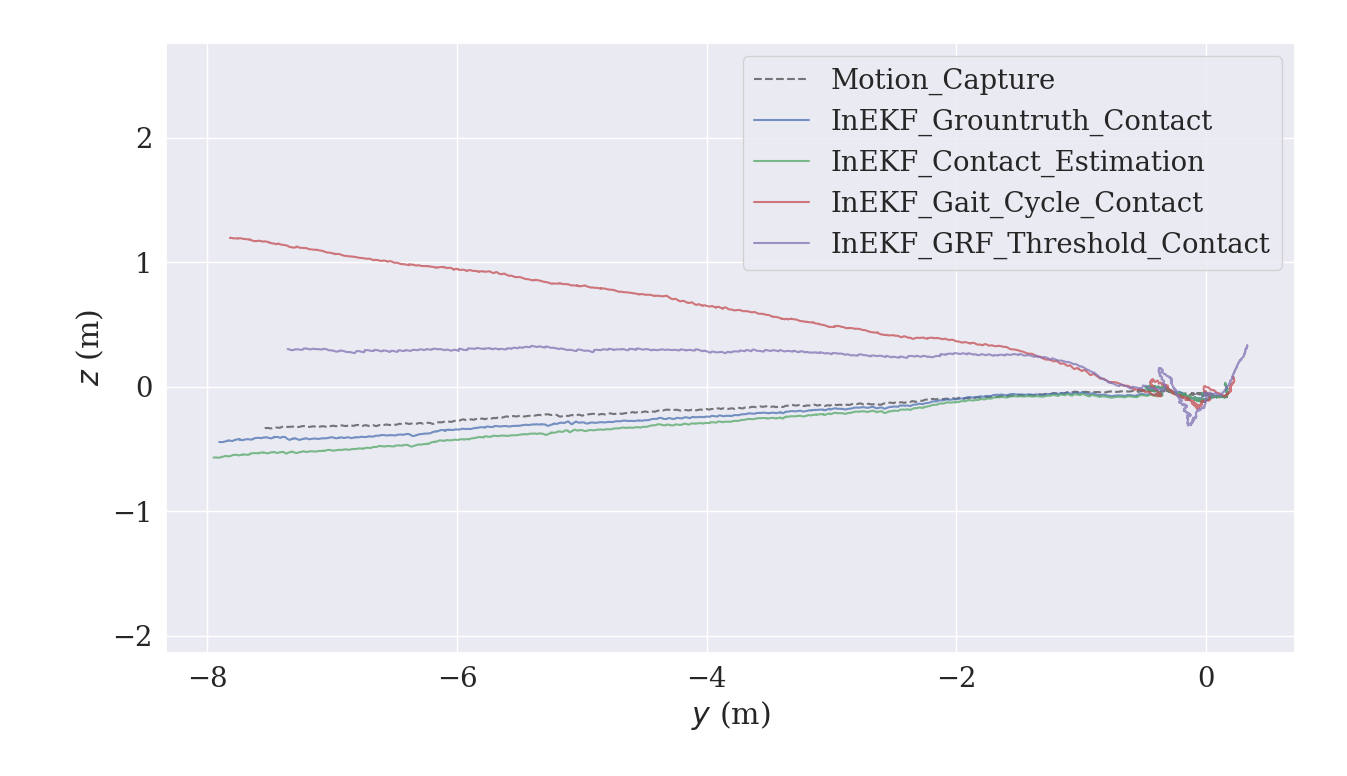}
    \caption{A side-view of the trajectories on the grass test sequence. The InEKF trajectory with gait cycle contact detection methods have a significant drift in the height (Z) axis.}
    \label{fig:mair_baseline_yz}
\end{figure}
\begin{figure}[t]
    \centering
    \includegraphics[width=0.99\columnwidth]{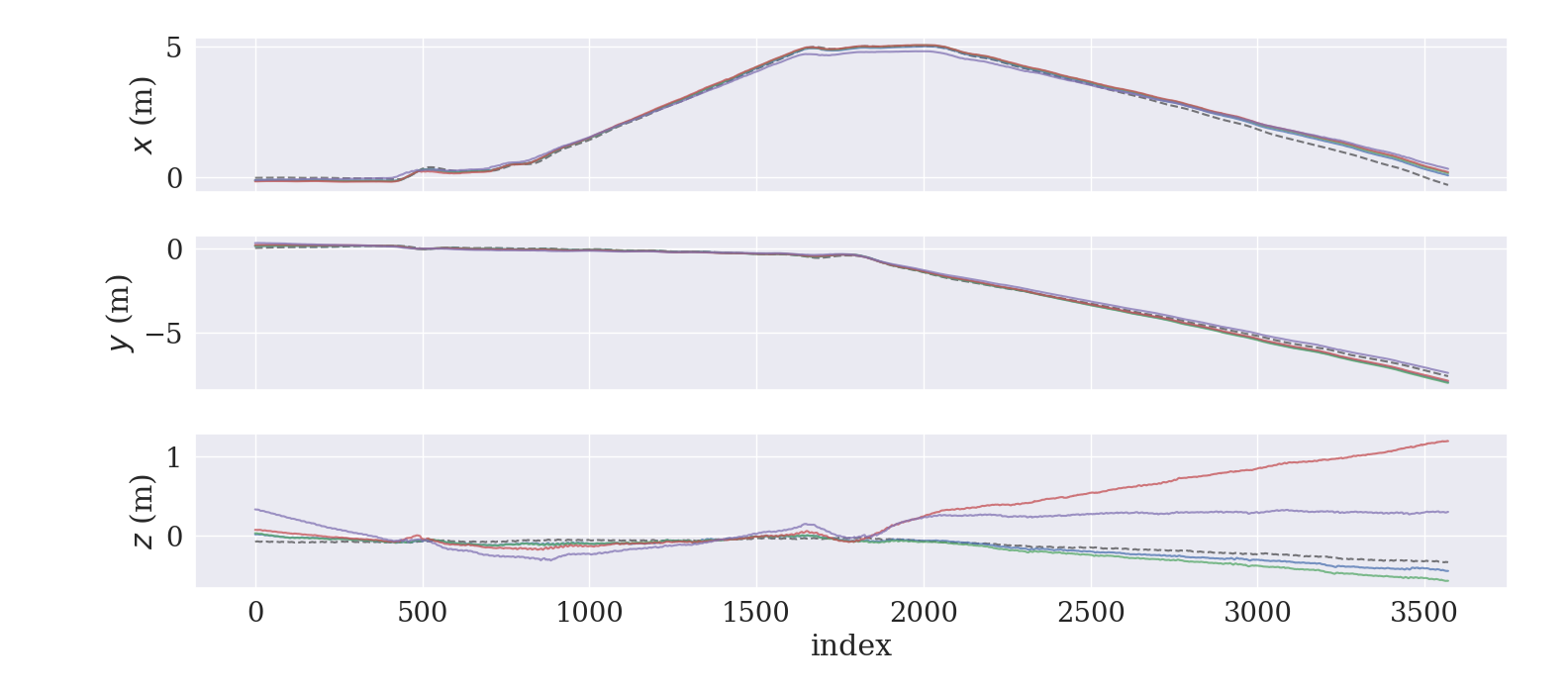}
    \caption{The section view of the trajectories. We can see that both gait cycle and GRF thresholding creates significant height (Z) drift in state estimation.}
    \label{fig:mair_baseline_section}
\end{figure}

\subsection{Comparison Against Other Contact Detection Methods}
Figure \ref{fig:mair_baseline_xy}, \ref{fig:mair_baseline_yz}, and \ref{fig:mair_baseline_section} show the trajectories of the InEKF with different contact estimation methods in the grass test sequence. We can see that the resulting trajectory from GRF thresholding gives larger drifts in the X-Y plane. The result with gait cycle method, although it shows a similar path to the proposed method in the X-Y plane, has a significant drift in the height (Z) axis, as shown in Figure \ref{fig:mair_baseline_yz} and \ref{fig:mair_baseline_section}.

\begin{figure}[t]
    \centering
    \includegraphics[width=0.99\columnwidth,trim={2.5cm 1cm 1cm 1cm},clip]{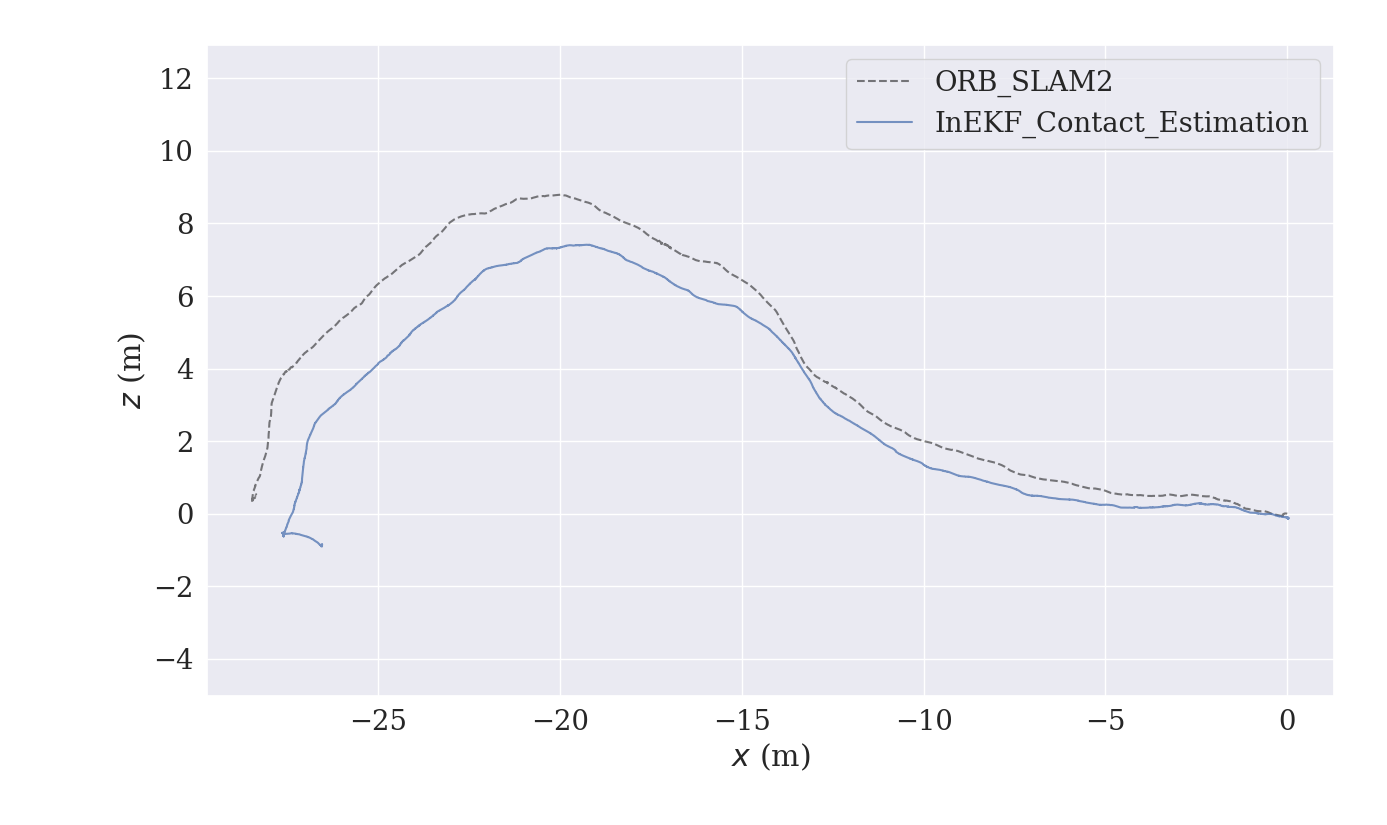}
    \caption{Trajectory of the forest test set. ORB SLAM2 fails close to the end of the sequence. This figure confirms the complementary role of the proposed proprioceptive state estimator. }
    \label{fig:forest_traj}
\end{figure}
\begin{figure}[t]
    \centering
    \includegraphics[width=0.99\columnwidth,trim={1cm 1cm 1cm 1cm},clip]{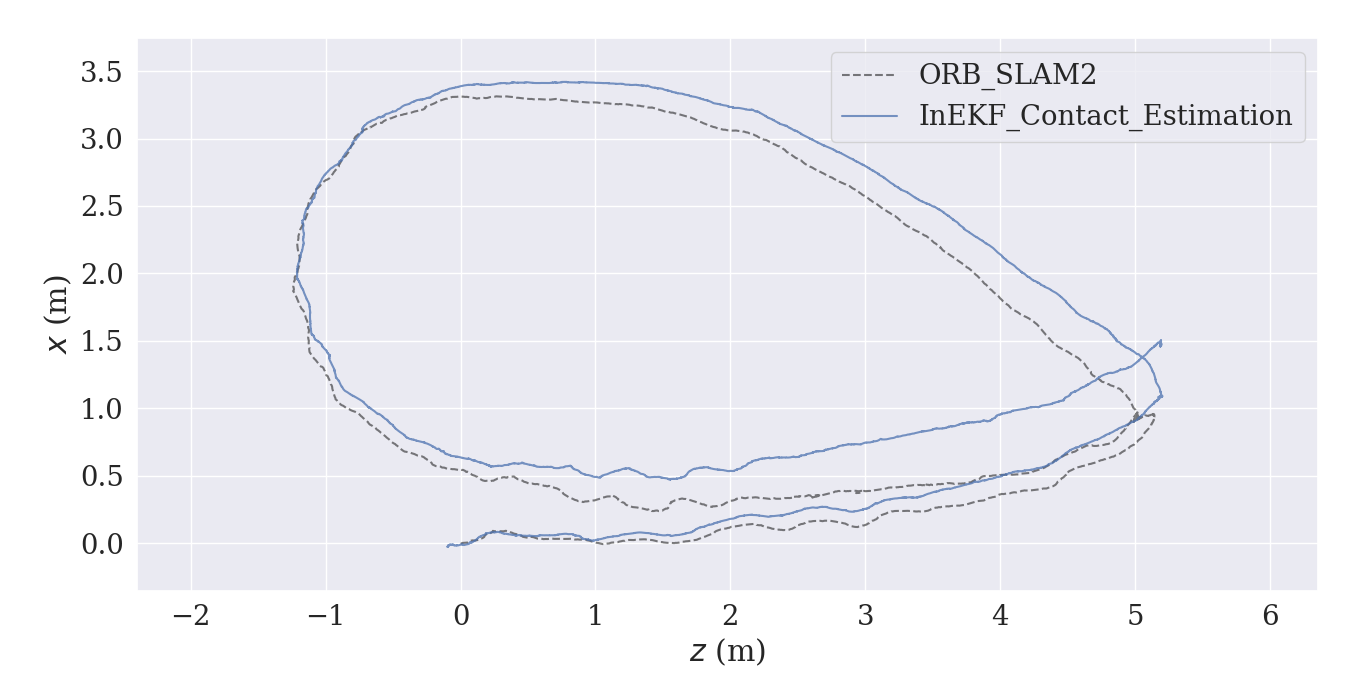}
    \caption{Trajectory of a test set with both small and middle pebbles. The robot walks in a circular shape starting from middle size pebbles to small size pebbles.}
    \label{fig:pebble_traj}
\end{figure}
\begin{figure}[t]
    \centering
    \includegraphics[width=0.99\columnwidth,trim={2.5cm 1cm 1cm 1cm},clip]{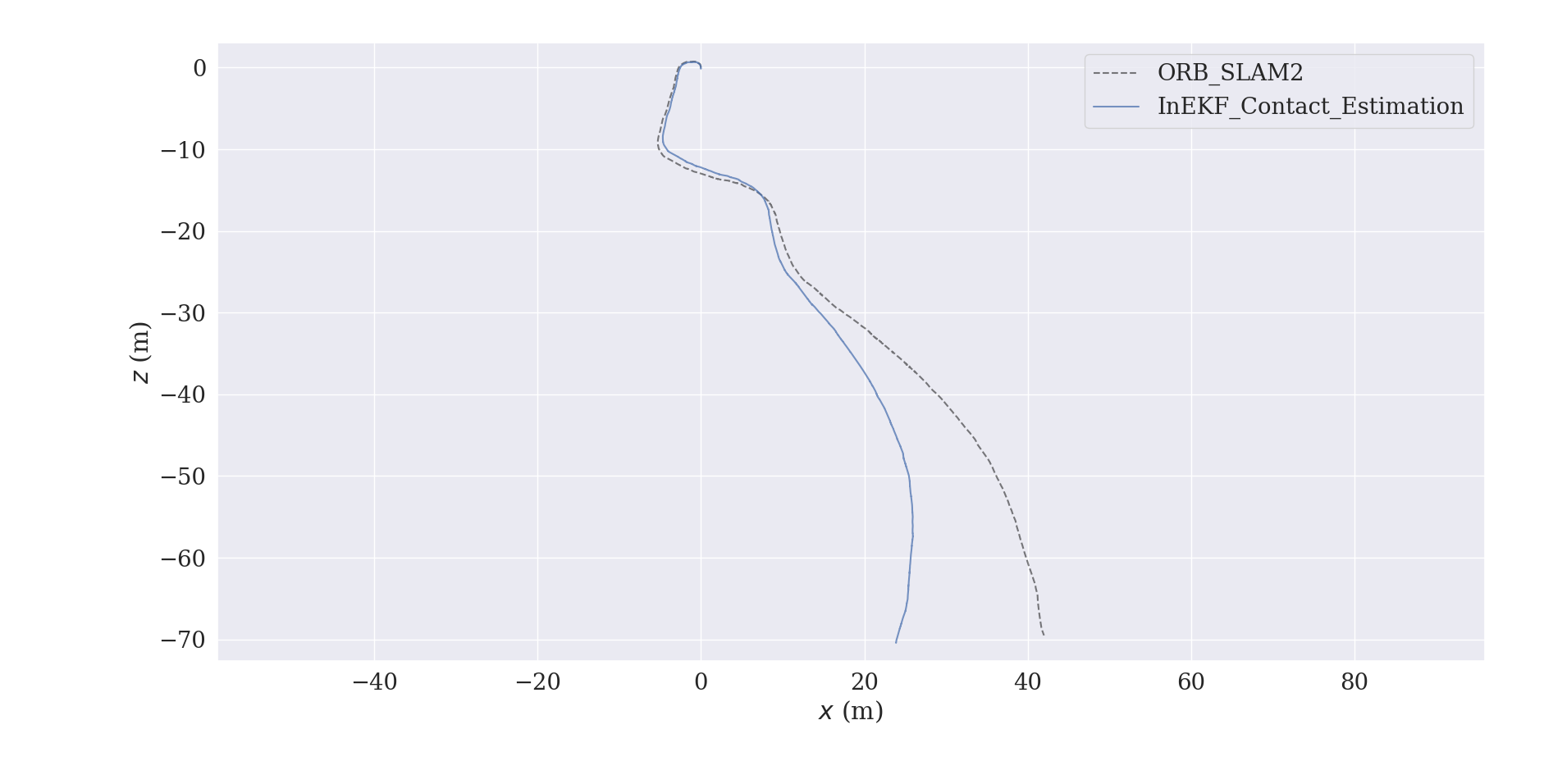}
    \caption{Trajectory of a test set with both sidewalk and asphalt road. the robot starts from the sidewalk, turns and walks into a sloping asphalt road. The total length of this path is around $100 \m$.}
    \label{fig:sidewalk_traj}
\end{figure}

\subsection{Extra Results on Contact-Aided Invariant Extended Kalman Filtering}
We present three extra trajectories on forest, small and middle pebbles, and sidewalk and asphalt road, as shown in Figure \ref{fig:forest_traj}, \ref{fig:pebble_traj}, and \ref{fig:sidewalk_traj}, respectively. These three test sets were not included in the training process. The forest test set consists of forest terrain described in the paper. In this experiment, ORB SLAM2 fails close to the end of the sequence due to a sudden sharp turn. In the pebbles test set, the robot walks in a circular shape starting from middle size pebbles to small size pebbles. As for the sidewalk and asphalt test set, the robot starts from the sidewalk, turns, and walks into a sloping asphalt road. The total length of the robot path in this sequence is around $100 \m$.

The three plots show that with the proposed contact estimator, the InEKF can successfully estimate the pose across 5 different terrains using only IMU and encoder measurements. It is worth noticing that the InEKF is an odometry-only method, that is, accumulating estimated poses without post-optimization or loop-closure detection. Thus, the error at each estimation is accumulated in the trajectory. The integration of this InEKF odometry into a real-time SLAM system is an interesting future work direction.

\end{document}